\documentclass[sn-mathphys]{sn-jnl}

\jyear{2022}%

\theoremstyle{thmstyleone}%
\theoremstyle{thmstyletwo}%

\theoremstyle{thmstylethree}%

\usepackage{enumitem}
\usepackage{graphicx}
\usepackage{subcaption}
\usepackage{placeins}
\usepackage{float}
\usepackage{longtable}
\usepackage{xcolor}
\usepackage{hyperref}
\usepackage{url} 
\usepackage{eucal}
\usepackage{multirow}

\newcommand{\Objets}{\ensuremath{\mathcal{O}}}
\begin{document}

\title[Introduction of a tree-based technique for real-time label retrieval]{Introduction of a tree-based technique for efficient and real-time label retrieval in the object tracking system}


\author*[1,3]{\fnm{Ala-Eddine} \sur{Benrazek}}\email{benrazek.alaeddine@univ-guelma.dz}

\author[2,3]{\fnm{Zineddine} \sur{Kouahla}}\email{kouahla.zineddine@univ-guelma.dz}

\author[2,3]{\fnm{Brahim} \sur{Farou}}\email{farou.brahim@univ-guelma.dz}

\author[2,3]{\fnm{Hamid} \sur{Seridi}}\email{seridi.hamid@univ-guelma.dz}

\author[2]{\fnm{Imane} \sur{Allele}}\email{imane.allele@gmail.com}

\affil*[1]{\orgdiv{Department of Computer Science}, \orgname{Ziane Achour University}, \orgaddress{\city{Djelfa}, \postcode{17000}, \country{Algeria}}}

\affil[2]{\orgdiv{Department of Computer Science}, \orgname{8 may 1945 University}, \orgaddress{\city{Guelma}, \postcode{24000},  \country{Algeria}}}

\affil[3]{\orgdiv{Labstic Laboratory}, \orgname{8 may 1945 University}, \orgaddress{\city{Guelma}, \postcode{24000},  \country{Algeria}}}


\abstract{This paper addresses the issue of the real-time tracking quality of moving objects in large-scale video surveillance systems. During the tracking process, the system assigns an identifier or label to each tracked object to distinguish it from other objects. In such a mission, it is essential to keep this identifier for the same objects, whatever the area, the time of their appearance, or the detecting camera. This is to conserve as much information about the tracking object as possible, decrease the number of ID switching (\texttt{ID-Sw}), and increase the quality of object tracking. To accomplish object labeling, a massive amount of data collected by the cameras must be searched to retrieve  the most similar (nearest neighbor) object identifier. Although this task is simple, it becomes very complex in large-scale video surveillance networks, where the data becomes very large. In this case, the label retrieval time increases significantly with this increase, which negatively affects the performance of the real-time tracking system. To avoid such problems, we propose a new solution to automatically label multiple objects for efficient real-time tracking using the indexing mechanism. This mechanism organizes the metadata of the objects extracted during the detection and tracking phase in an Adaptive BCCF-tree. The main advantage of this structure is (1) its ability to index massive metadata generated by multi-cameras, (2) its logarithmic search complexity, which implicitly reduces the search response time, and (3) its quality of research results, which ensure coherent labeling of the tracked objects. The system load is distributed through a new Internet of Video Things infrastructure-based architecture to improve data processing and real-time object tracking performance. The experimental evaluation was conducted on a publicly available dataset generated by multi-camera containing different crowd activities. The experimental results showed excellent and promising performance.}


\maketitle

\section{Introduction}
Today in smart cities, video surveillance systems (VSS) have become an essential part of its infrastructure. These systems play an indispensable role in our lives due to their enormous benefits, like making public and private places safe and improving safety in our community. VSS consists of placing cameras in the surveillance environment, tracking the people observed, and detecting suspicious behaviors through an intelligent system that decides based on the analysis of the sequence of scenes.

\begin{figure}[h]
    \centering
    \includegraphics[width=0.7\textwidth]{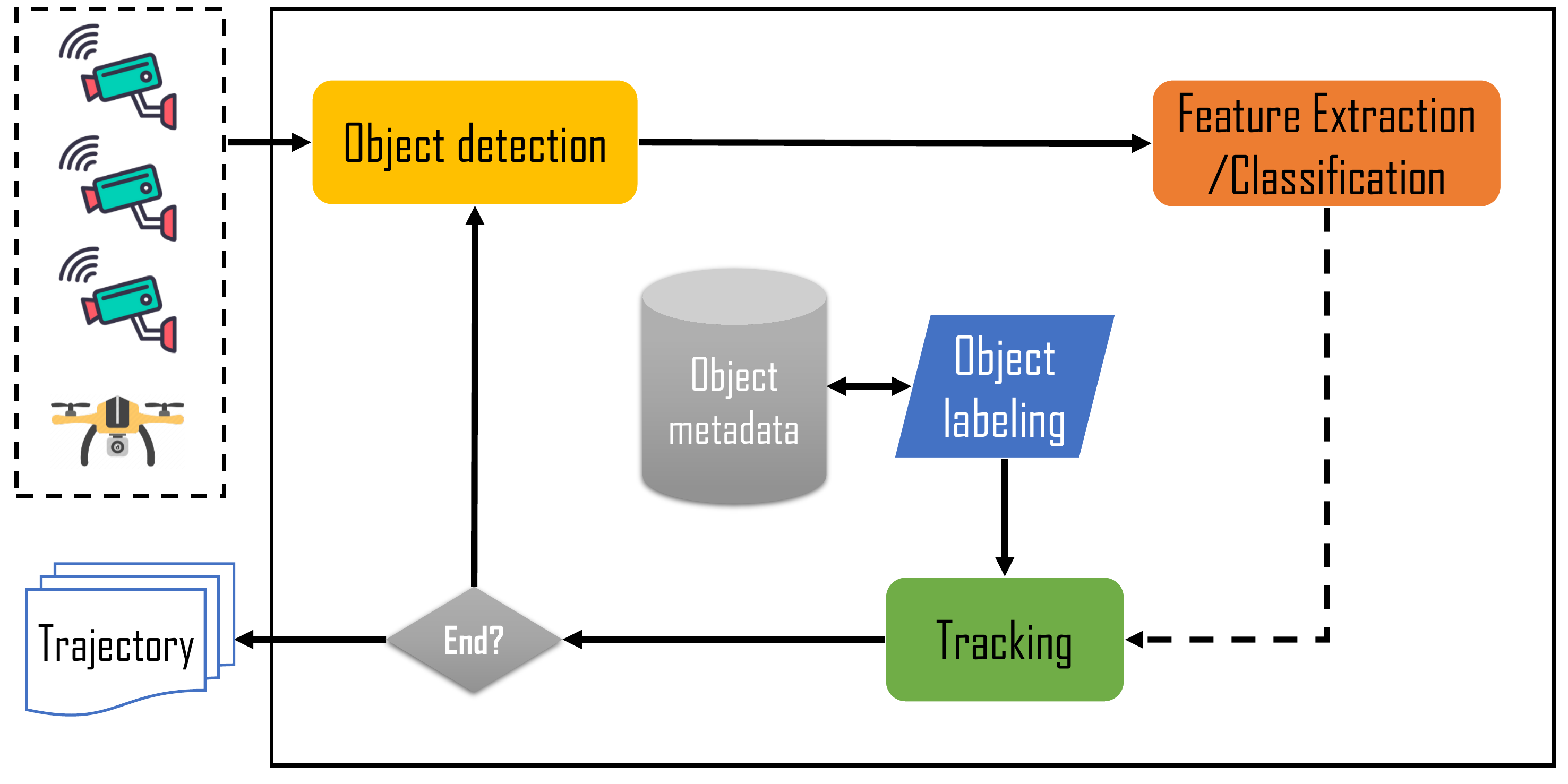}
    \caption{Main processing models of the tracking system}
    \label{fig:Processingmodel}
\end{figure}

Tracking moving objects is a fundamental problem in computer vision. Many applications make it important, such as surveillance, human motion analysis, traffic monitoring, and human-computer interfaces. In short, tracking moving objects is the automatic correspondence between the objects seen in the current frame and those in the previous frame.
The main processing models of this system are illustrated in the flow chart in Figure \ref{fig:Processingmodel}.

The task of object labeling is essential to distinguish the identities of tracked objects from each other and conserve a maximum of information on the tracked objects. To track and extract the trajectory of multi objects, the unique label assigned to each object must be maintained for all the tracked objects, whatever the area, the time of their appearance, or the detecting camera. Changing or switching this label can lead to :

\begin{enumerate}
    \item Increase the number of identity changes (\texttt{ID-Sw})
    \item Increase the rate of trajectory fragmentation (\texttt{\#Frag}).
    \item Decrease the rate of objects that have been tracked by at least 80\% (the mostly tracked -\texttt{MT}-) of their lifespan.
    \item Increase the rate of objects that have been tracked at most 20\% (the mostly lost -\texttt{ML}-) of their lifespan.
\end{enumerate}

As a result, the quality of tracking is also decreased.

In the object labeling phase, each object sample $O$ is represented by its meta-data, denoted $m_d$, and label, denoted $lab$. Then, the similarity function $\Psi(\cdot,\cdot)$ is applied between $m_d$ and labeled metadata in the dataset ($D_s$). The label $lab_x$ assigned to the detected object returns to the object with the closest distance (nearest neighbor).

\[O\left(m_d,lab_x\right)=\ argmin\{\ \Psi\left(m_d,x_d\right),\ \forall O_x(x_d,\ lab_x)\ \in D_s\}\]

To accomplish this process, usually, the system performs a sequential search on all the metadata without considering the time needed to retrieve the most similar object. The problem here resides in the linear complexity ($O(n)$) of the sequential search to retrieve the objects labels in data increasing with time, the number of objects, and the number of cameras in the network. This increase makes the search time much longer. Thus, the system can find the label after the detected object has left the detector camera field of view (FoV). This problem appears especially in large-scale real-time VSS networks, where metadata becomes 'big metadata'. In this case, the organization of metadata for quick access is mandatory.

In this context, efficient indexing and searching in massive data is one of the most promising paradigms for addressing our issues. Indexing is a data organization step that should allow efficient access to data when performing similarity queries. Indexing techniques aim to build a data structure that organizes the database to provide quick access to the objects in a database by reducing the search space, the cost of input-output, and the complexity. In other words, the index provides the efficient implementation of associative search \cite{gaede1998multidimensional}.

Among the indexing techniques in the literature, there are tree-based indexing structures that have gained popularity in recent years \cite{Khettabi2022Clustering, kouahla2022survey, Allele2021Automatic, Kemouguette2021Cost, moriyama2021vd, benrazek2020efficient}. These structures are dynamic, do not require periodic reorganizations, and ensures continuous indexing. Thanks to the logarithmic complexity of the insertion and search provided by tree structures, the search time is reduced logarithmically depending on the number of indexed objects. For this reason, our proposed solution will be based on tree indexing structures to leverage its advantages.


As we mentioned before, we focus on the large-scale real-time VSS that has become ubiquitous. The main challenge facing these systems is the rapidly growing number of surveillance cameras \cite{sultana2019choice}. In 2017, Anup et al. in \cite{mohan2017internet} studied the increase in the number of cameras connected to the Internet and noted that by 2030, there would be 13 billion cameras connected. With this increase, the amount of visual data generated by these cameras has exploded. This expansion has created the need to store, serve, and process data and introduced several constraints like high latency, high bandwidth, and high-energy consumption \cite{benrazek2019efficient, benrazek2019ascending}. To address these challenges, researchers have focused on using the modern computing paradigm of the IoT. As a result, a new generation of VSS has emerged, called the "Internet of Video Thing" or simply "IoVT" \cite{sultana2019choice, mohan2017internet, sammoud2017real, chen2020internet}. IoVT is a part of the IoT that can efficiently process large volumes of visual data \cite{sultana2019choice}. Compared to the previous generation, the IoVT framework provides multiple layers based on modern computing paradigms such as mist, edge, fog, and cloud computing as a networking infrastructure that supports immediate decision-making due to bringing services to the edge through decentralization of cloud services \cite{sultana2019choice}. IoVT aims to develop a more efficient, flexible, and cost-effective VSS, adapted to the new requirements of smart cities in terms of citizen security in public and private places. We expect the IoVT paradigm to improve VSS performance further as it becomes more powerful and operates more efficiently in real-time.

This article is intended to address the following issues:

\begin{itemize}
    \item The need to design a flexible, robust, and self-adaptive platform supports a large-scale VSS network by leveraging the existing modern infrastructure. To address this issue, we propose a new architecture of a VSS based on IoVT infrastructure.

    \item The need for new solutions to increase the quality of tracking and the operation of the tracking system in real time due to the increasing complexity of the system's tasks, ensuring the immediate processing of actions or decisions taken by the system entities. To mitigate this issue, we propose a new solution to automatically label multiple objects for efficient real-time tracking based on the indexing mechanism. Our solution aims to organize the metadata of the extracted objects during the detection and tracking phase into an Adaptive BCCF-tree. BCCF-tree is a recent indexing structure that has proven its efficiency in big IoT data \cite{benrazek2020efficient, Allele2021Automatic}. The main advantages of this structure are:

          \begin{enumerate}
              \item Its ability to index massive metadata generated by multi-cameras,
              \item Its logarithmic search complexity, which implicitly reduces the search response time, then supports the real-time performance, and
              \item Its research quality results, which improve the quality of the labeling process, then the quality of tracking.
          \end{enumerate}
\end{itemize}
The experimental evaluation was conducted on a publicly available dataset generated by muli-cameras containing different crowd activities. The experimental results were excellent and promising.

The present work is structured as follows. Section \ref{section2} provides a brief overview of label search for object tracking in VSS. Then, Section \ref{sec:architecture} presents the proposed architecture of our VSS's proposed architecture based on the IoVT infrastructure. Section \ref{section3} discusses the functionality and workflows of the proposed automatic labeling-based tracking system. Next, Section \ref{section4} highlights the improvements provided by the Adaptive BCCF-tree for automatic object labeling. The results are discussed in Section \ref{section5}, while Section \ref{section6} concludes the paper.

\section{RELATED WORK }\label{section2}

In VSS, label retrieval using a traditional DBMS (Database Management Systems) is not adequate for real-time operation on high-volume, continuous, and time-varying data streams. Real-time search requirements are different from conventional applications and need new solutions to address the long-running search problem to produce continuous and incremental results.

In \cite{huang2009moving, huang2011bayesian}, Huang and Wang propose a new technique for labeling and mapping objects on multi-camera systems. The proposed method operates in two stages. The first step is to collect and merge detection information from several cameras to develop the scene knowledge. Based on this letter, the objects and their positions in a 3D scene are identified thanks to the combination of Markov Chain Monte Carlo (MCMC) and Mean-Shift clustering. Then, a probabilistic method based on the Markov network is used on the 3D scene to label the objects. The method remains complex and unsuitable for real-time performance despite the results obtained, especially in large-scale networks. Moreover, the proposed technique focuses on object labeling only to correspond objects detected by overlapping multi-cameras simultaneously and does not consider the problem of object labeling during the long-term tracking process and non-overlapping cameras.

In the same context, Khan and Shah \cite{khan2003consistent}, and Simone et al. \cite{calderara2005consistent} presented a system based on the FoV lines of the camera to hand off labels from one camera to another (identify common objects). The FoV information was learned during a training phase. Using this information, when an object was viewed in one camera, all the other cameras in which the object was visible could be predicted. In addition to the previous limits cited for \cite{huang2009moving, huang2011bayesian}, this technique requires considerable overlap in FoVs of cameras, which require prohibitive cost and computational resources for surveillance of large-scale areas. Moreover, this technique requires the visibility and appearance of human feet, which may not be available in large-scale surveillance networks in a crowd.

In \cite{huang2009ips}, an inter-camera handoff communication control is proposed to assist client applications in changing monitoring connections automatically when a tracked object moves among multiple cameras in a large-scale surveillance environment. Among the processes of this system is the labeling of tracked objects compared with previous metadata. This process is performed using a SQL server based on object metadata organized into database tables. As we mentioned at the beginning of this section, this technique is obsolete because of its failure for real-time performance on high-volume, continuous and time-varying data streams.

Using only a single label to match all tracking fragments or tracklets of the same target in long-term tracking is challenging. For this reason, Tianyi et al. in \cite{liang2021generic} propose post-processing of the tracking fragment or tracklets. The proposed clustering method merges labels hierarchically under the same label from different tracklets based on the cosine distance between the metadata. The main idea is to reuse the tracklets and link them to obtain more intact trajectories. Unfortunately, this technique is not suitable for real-time performance due to the need for tracklets (tracking results). In addition, this technique relies on single object tracking, which makes it very prohibitive and requires more time when tracking multiple objects, especially in large-scale multi-cameras.

Jia et al. \cite{liu2009automatic, tong2011automatic} propose a new solution to detect multiple players, label them, and efficiently tracking them automatically in broadcast soccer videos. In the player labeling phase, the authors represent each player sample by its metadata (represented as a histogram). To retrieve the corresponding label of the players, the authors perform a sequential search between the metadata of the player in question and all the metadata of each sub-model. The label assigned to the player corresponds to the label of the sub-model that has the nearest Bhattacharyya distance (nearest neighbor). Even though the method is generally effective, it should be considered that this technique has been applied to a limited number of objects (number of players in the game) and focuses on a single camera. Due to the sequence search, the method is unsuitable for real-time scenarios, especially if applied in a context outside of soccer video broadcasting where the number of objects and cameras are more significant. In this case, we fall into the linear search time problem.

Based on this study, and despite the lack of work in this area in the literature, we find that the problem of automatic object identification or labeling has not been effectively addressed in real time and with high tracking quality, especially in large-scale networks.

\section{SYSTEM ARCHITECTURE OVERVIEW}
\label{sec:architecture}

\begin{figure}
    \centering
    \includegraphics[width=0.8\textwidth]{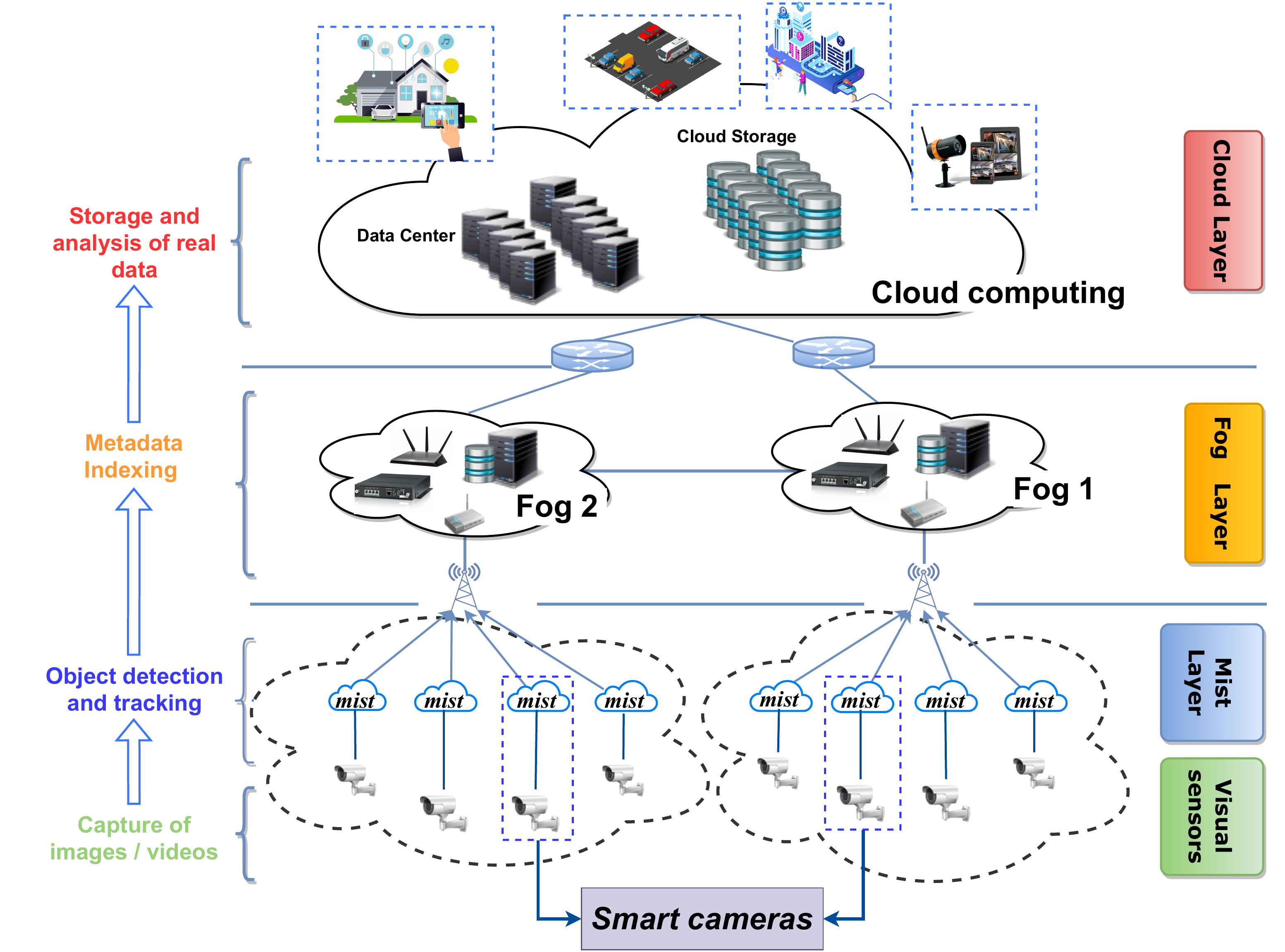}
    \caption{Video surveillance system data indexing and discovery in an IoVT-based environment}
    \label{fig:system overview}
\end{figure}

VSS are one of the most significant visual data sources in the real world these days. With recent developments, distributed VSS has become ubiquitous due to the increasing demands for large-scale security in all areas, especially in smart cities. In these applications, distributed VSS needs to deal with the fast-growing number of surveillance nodes which impose several requirements, like high latency, high bandwidth, high energy consumption, processing power, and storage capacity. To meet these requirements, the IoVT infrastructure can be a promised solution. This infrastructure adapts and leverages modern computing paradigms (cloud, fog, mist, etc.) to meet the above requirements and improve quality of service.


This section introduces our distributed VSS's architecture based on the IoVT computing paradigm to improve real-time object tracking quality. As shown in Figure \ref{fig:system overview}, the proposed system is distributed over four layers. Each layer has specific tasks among the tasks for real-time tracking of moving objects. These tasks are assigned according to the task requirements and the resources available in the layer.

The bottom layer consists of smart cameras. These cameras have processing, storage, and communication capabilities due to integrating modern micro-controller technologies. From an architectural point of view, this type of camera is represented in two layers: (i) the visual sensor layer and (ii) the mist computing layer. The role of the visual sensor layer is to capture the events located in their FoV. The mist layer pushes processing even further to the network edge, decreasing latency and increasing subsystems' autonomy. In our context, the first task assigned to each mist node is detecting objects from the incoming frames. Then, it extracts each detected object's metadata and sends them to the next layer to identify or label them. Then, the mist node starts the tracking processes until the tracked object leaves the camera's FoV. At the end of the tracking, the mist node sends the tracked objects' metadata to the Fog layer.

\begin{figure}
    \centering
    \includegraphics[width=0.5\textwidth]{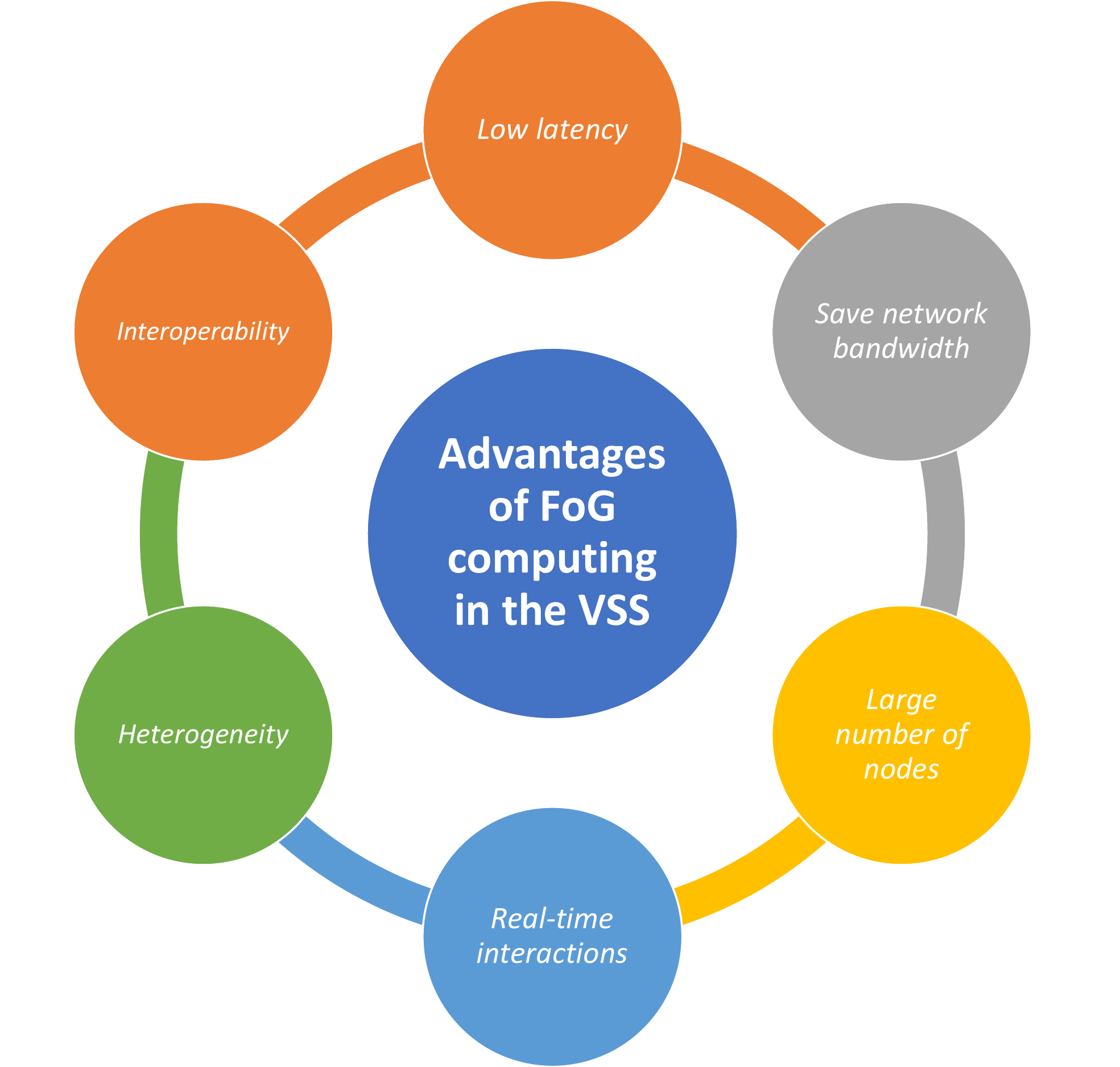}
    \caption{Advantages of fog computing in the video surveillance system}
    \label{fig:FoginVSS}
\end{figure}

The fog layer has many advantages for the VSS, which prompts us to integrate it into our system. Figure \ref{fig:FoginVSS} briefly shows the most important. Each fog node manages and controls a set of smart cameras located in their geographical area. Each fog node receives data from all nodes in the lower layer (mist layer) and aggregates it. Then, the fog nodes will index this data in the proposed indexing structure to identify (labeling) the requested objects and be shared with other cameras or neighboring fog nodes if necessary. Our indexing structure is created in this layer to develop an efficient real-time distributed indexing mechanism that allows us to search in real-time. Thus, each fog node has its own structure known as a local structure. The real or raw data (images/videos) indexing in the fog layer is transmitted to the cloud layer to store it.

The fourth layer is the cloud computing layer. The cloud or data-center is adopted for the following reasons: (i) the large storage capacity to save the raw system data transferred via the lower layer, and (ii) the computing power to execute the robust algorithms like analyzing objects' behavior.

\section{Proposed automatic labeling system}
\label{section3}

\begin{figure}[h]
    \centering
    \includegraphics[width=0.9\textwidth]{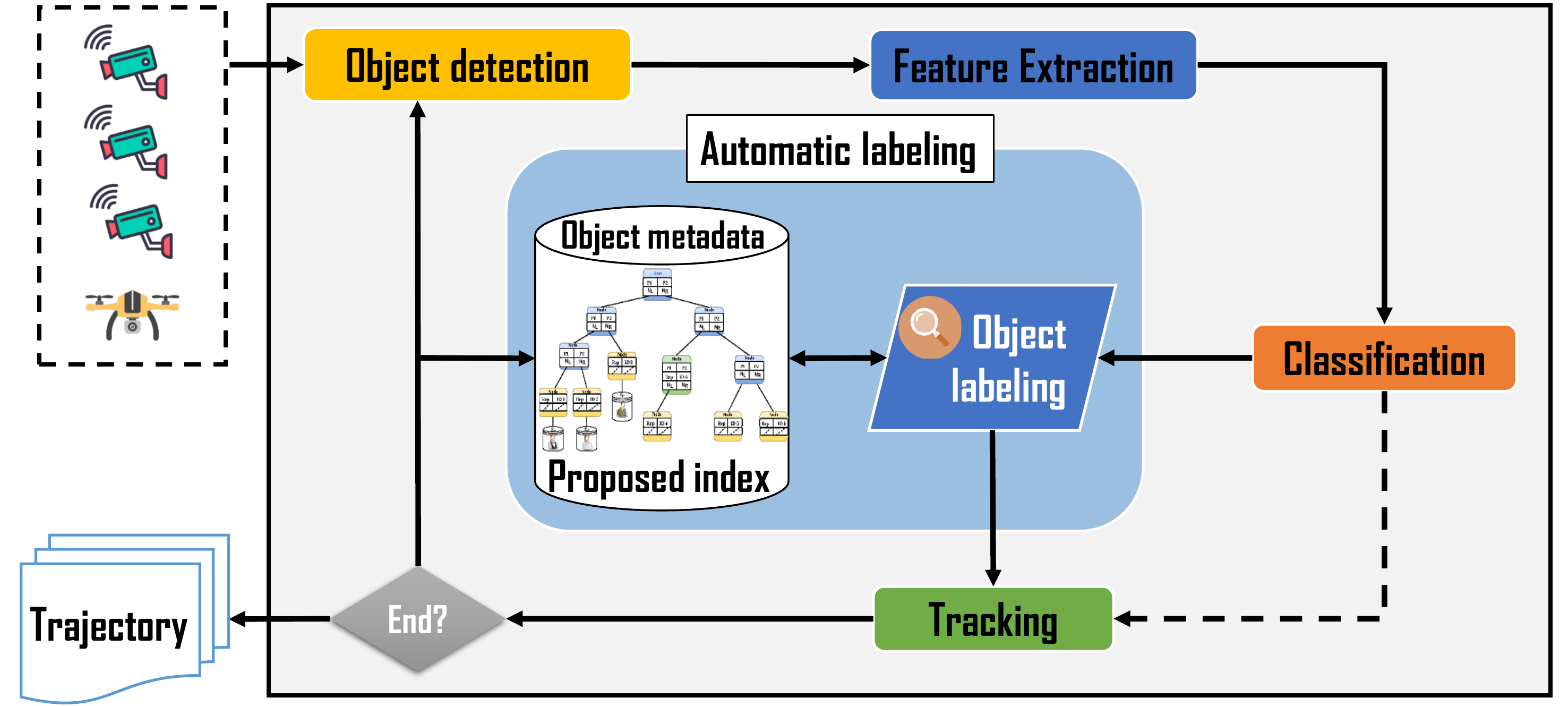}
    \caption{Proposed tracking system based on automatic labeling}
    \label{fig:ProposedSystem}
\end{figure}

The main objective of our proposed system is to improve the quality and real-time performance of object tracking by introducing a new automatic tracking object labeling system based on a tree-based indexing structure that replaces the sequential search process of conventional systems (see Figure \ref{fig:Processingmodel}), as shown in Figure \ref{fig:ProposedSystem}.

\begin{figure}[h]
    \centering
    \includegraphics[width=0.8\textwidth]{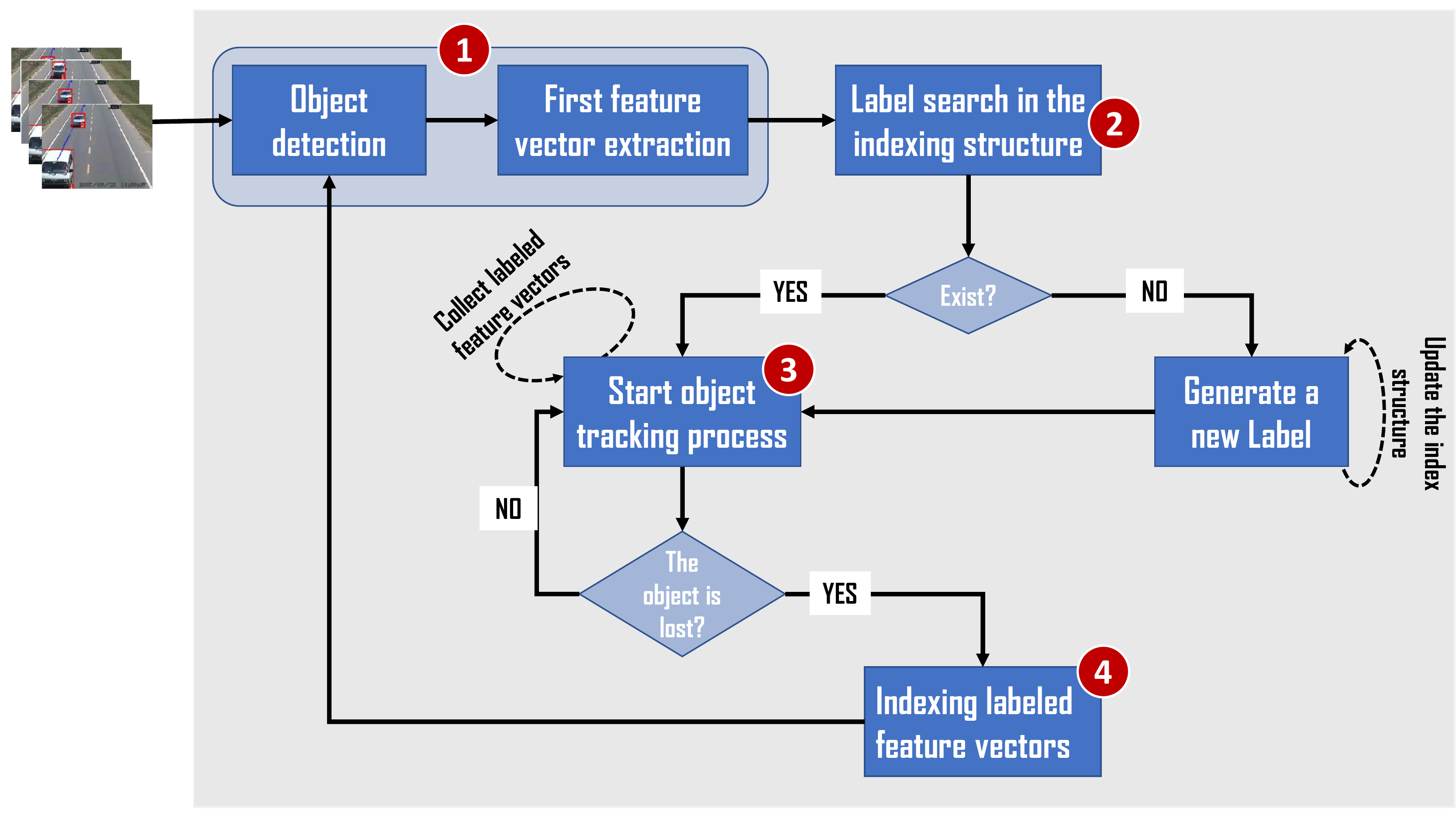}
    \caption{System operation diagram}
    \label{fig:diagram}
\end{figure}

Our contribution in this paper focuses on finding an alternative to the conventional labeling system shown in Figure \ref{fig:Processingmodel}, which is not suitable for large-scale networks and real-time operation. And the suggested solution is to propose a new automatic object labeling system based on a tree indexing technique. For this reason, we use Simple Online and Real-time Tracking (SORT) \cite{wojke2017simple} as a multiple objects tracking system, and we introduce our solution into it. The main steps of this system are shown in Figure \ref{fig:diagram}.

\subsection{Detection and first vector extraction} \label{step1}

Object detection is one of the main tasks of our system. This task consists of determining the location of objects on the frames recorded by the cameras in the surveillance area and classifying these objects (humans, animals, vehicles, etc.) to keep only the information of the objects relevant for the surveillance. Several approaches have been proposed in the literature. Among these approaches, we are interested in using YOLO (You Only Look Once)\cite{redmon2016you} a real-time object detection neural network  as an object detector. This approach has gained popularity in recent years due to its superior performance compared to other object detection techniques, fast speed, smaller memory footprint, etc \cite{huang2020design, bochkovskiy2020yolov4}.

For feature vector extraction, the system uses a CNN architecture trained on a large-scale person re-identification dataset that contains over 1,100,000 images of 1,261 pedestrians \cite{zheng2016mars}. The CNN architecture consists of a large residual network with two conventional layers followed by six residual blocks. The global feature map of dimension 128 is computed in dense layer 10 \cite{wojke2017simple}. The CNN network takes as input a BGR color image and a matrix of bounding boxes in $(x, y, w, h)$ format, and returns a matrix of corresponding feature vectors \cite{wojke2017simple}.

\subsection{Label search} \label{step2}

This step consists in finding the label of the detected object from its feature vector (metadata) extracted in the previous step (Subsection \ref{step1}) on previously labeled metadata according to the similarity metric. More precisely, the detected object receives the label of the most similar object (nearest neighbor) based on a previously defined similarity threshold. Usually, in conventional system (Figure \ref{fig:Processingmodel}), to retrieve the adequate label, a sequential search is performed on all the metadata without considering the time needed to find the most similar object. Despite the simplicity of this process, the big problem resides in large-scale VSS where the data increases dramatically over time, with the number of objects tracked and the number of active cameras in the network producing a large volume of data. This increase increases the search time, which negatively influences the real-time tracking process and reduces their efficiency.

As a solution to this linear complexity, we propose to use a tree-based indexing structure, which provides a faster search than sequential search with logarithmic complexity ($log_b(n)$) instead of linear complexity ($O(n)$). Based on this solution, the label search step is performed with a 1-$nn$ search query (sometimes called "exact search") using the first feature vector in the suggested indexing structure, which contains previously labeled objects. The suggested structure is discussed in the next section (Section \ref{section4}).

\subsection{Object tracking} \label{step3}

This step makes it possible to follow the trajectory of an object detected in a well-defined period by locating its positions at each moment. After identifying or labeling the object --in Subsection \ref{step2}--, the tracking process will be triggered to collect a maximum of information on the object of interest. The model used in our system for tracking is the standard Kalman filter with constant velocity motion model, in which we take the bounding box coordinates $(u, v, \gamma, h)$ as direct observations of the object state. Where, ($u, v$) means the center of the bounding box, $\gamma$ is the ratio with width to height, and $h$ is the height of the bounding box.

The standard Kalman filter consists of two steps:

\begin{itemize}
    \item \textit{Prediction step}: allows predicting the state of the object in the future thanks to its dynamic model.
    \item \textit{Correction/ Update step}: corrects the observation model such that the error covariance of the estimator is minimized.
\end{itemize}

\subsection{Labeled metadata indexing}\label{step4}

This step consists in inserting and indexing the set of metadata (labeled feature vectors) collected during the object tracking (Subsection \ref{step3}) in the suggested indexing tree structure. The objective of this structure in this step is to keep as much as possible the metadata collection of the same object in the same place and avoiding their dispersion to improve the search time and the quality of the identification of the objects in real-time.

\section{Index structure used for label search and indexing}
\label{section4}

This section presents the indexing structure used to better organize the metadata and provide quick access to this data, reducing the search space and search time. Among the recently proposed tree indexes, the BCCF-tree \cite{benrazek2020efficient}. The BCCF-tree is an efficient indexing structure for indexing massive IoT data. This structure is based on recursive partitioning of the space through the k-means clustering algorithm that efficiently separates the space into non-overlapping subspaces to improve the quality of research and discovery. Despite the efficiency of BCCF-tree in terms of search costs, it still suffers from the complexity of its construction due to the convergence cost of the k-means algorithm. Furthermore, this structure is based on the incremental construction, which is done by a recursive path with a descending phase from the root node followed by an ascending phase.

    {\setlength{\arrayrulewidth}{0.5mm}
        \renewcommand{\arraystretch}{1.2}

        \begin{table}
            \caption{ Notations, functions and their meaning}
            \label{tab:notations}
            \centering
            \begin{tabular}{c p{7cm}}

                \hline
                Notation                                   & Description                                                                                                                                                                                \\ \hline
                $p_1$ \& $p_2$                             & two pivots, with $d(p_1, p_2) > 0$                                                                                                                                                         \\
                $N_L$ \& $N_R$                             & left and right sub-tree of $N$                                                                                                                                                             \\
                $\mathcal{O}$                              & a set of object metadata                                                                                                                                                                   \\
                $o$                                        & a sample of object metadata, with $o \in \mathcal{O}$                                                                                                                                      \\
                $o_q$                                      & metadata of the requested object                                                                                                                                                           \\
                $\mathcal{N}$                              & the nodes of the Adaptive BCCF-tree                                                                                                                                                        \\
                $lab_i$                                    & a profile label of the object $o_i$                                                                                                                                                        \\
                $rep_i$                                    & a representative feature vector of the object $o_i$                                                                                                                                        \\
                $c_{max}$                                  & the maximum cardinal assigned to a container of a leaf node, with $1 \leq c_{max} \leq \|E\|$                                                                                              \\
                $E$                                        & a subset of the object's metadata that will be indexed ($E \subseteq \mathcal{O}$)                                                                                                         \\
                $E_c$                                      & a container or bucket consisting of a subset of the indexed objects' metadata, with, $\|E_c\| \leq c_{max} $                                                                               \\

                $\beta$ \& $\zeta  $                       & two predefined similarity thresholds                                                                                                                                                       \\
                \hline Functions                           & Description                                                                                                                                                                                \\ \hline
                $d(\cdot, \cdot)$                          & distance function from a pair of elements of $\mathcal{O}$ to a positive or zero real ($d: \mathcal{O} \times \mathcal{O}\to \mathbb{R}^+$)                                                \\
                $\texttt{New-lab}(\cdots)$                 & a function returns a new and unique label for a given object $o_x$                                                                                                                         \\
                $\texttt{New-node-leaf}(\cdots) $          & \multirow{3}{*}{\begin{minipage}{7cm}three functions to create the different nodes of a BCCF-tree (internal node, internal profile node, and leaf profile node) as shown in Figure \ref{fig:BCCF*}. \end{minipage}}                                                                                                                                                 \\
                $\texttt{New-node-profile-leaf}(\cdots)$   &                                                                                                                                                                                            \\
                $\texttt{New-node-internal}(\cdots)$       &                                                                                                                                                                                            \\
                $\texttt{Pointerswitch}(\cdots)$           & a function allowing to switch the pointers of two nodes of the tree.                                                                                                                       \\
                $\texttt{Partition}(\cdots)$               & a function allowing to split a set of metadata into two groups, whith :                                                                                                                    \\

                                                           & $
                    \stackrel{\Delta}{=}
                    \left\{
                    \begin{array}{ll}
                        G_1 = \{o \in \mathcal{O}, d(o, p_1) \le d(o, p_2) \} \\ & \\
                        G_2 = \{o \in \mathcal{O}, d(o, p_1) > d(o, p_2) \}
                    \end{array}
                \right.$                                                                                                                                                                                                                                \\
                $\texttt{mean}(\cdots)$                    & a a function to compute the average of two metadata. If we consider that these two metadata are two points in space, then the $mean$ corresponds to the center of gravity of these points. \\

                $\texttt{ChangeToInternalprofile}(\cdots)$ & a function allowing to switch a leaf node to an internal profile node, respecting the data structures presented in Figure \ref{fig:TypeNode}.                                              \\ \hline
            \end{tabular}

        \end{table}
    }

According to the previous section, labeled metadata indexing (Subsection \ref{step4}) is the insertion of a set of the same object metadata into the index structure. To avoid (over)-partitioning of this data and keep as much as possible similar data in the same place in order to reduce the search space, we avoided the use of incremental insertion and suggested the use of the batch insertion mechanism as a solution. This mechanism provides a low insertion cost for each new metadata set received without degrading query performance. For this purpose, we adapted the BCCF-tree structure to our automatic labeling system. The Adaptive BCCF-tree is shown in Figure \ref{fig:BCCF*} and consists of three types of nodes, namely:

\begin{figure}
    \centering
    \includegraphics[width=0.8\textwidth]{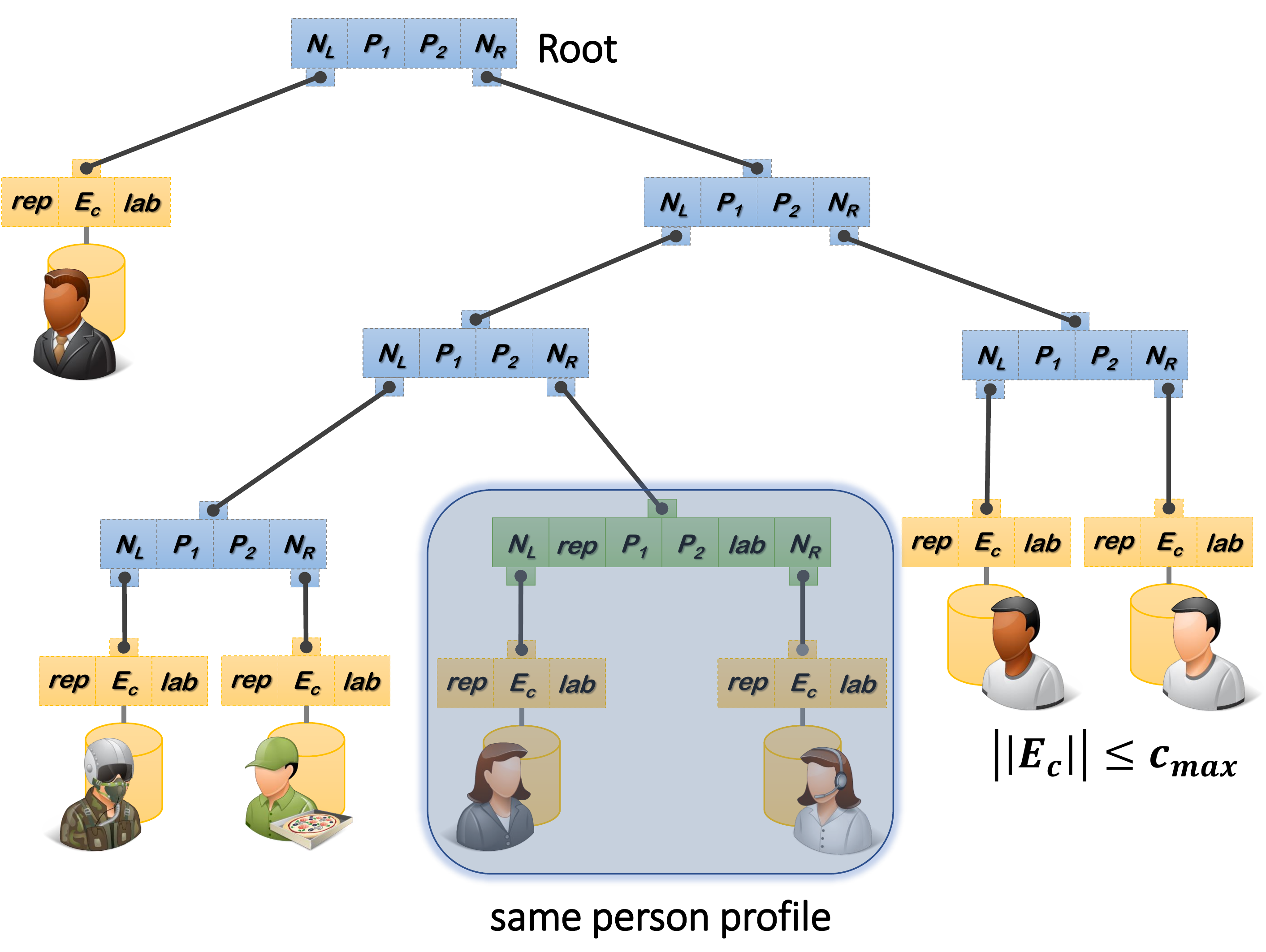}
    \caption{Adaptive BCCF-tree for automatic object labeling}
    \label{fig:BCCF*}
\end{figure}

\begin{figure}
    \centering
    \begin{subfigure}[b]{0.3\textwidth}
        \centering
        \includegraphics[width=\textwidth]{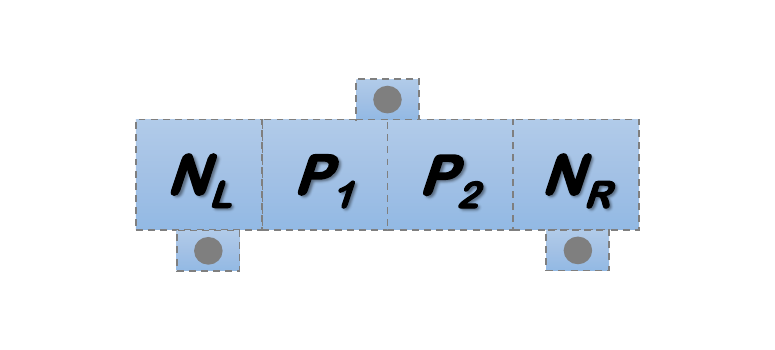}
        \caption{Internal node}
        \label{fig:Internalnode}
    \end{subfigure}
    \hfill
    \begin{subfigure}[b]{0.3\textwidth}
        \centering
        \includegraphics[width=\textwidth]{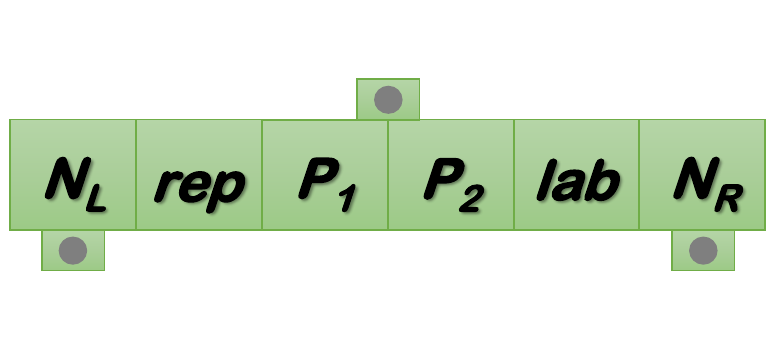}
        \caption{Internal profile node}
        \label{fig:Internalprofilenode}
    \end{subfigure}
    \hfill
    \begin{subfigure}[b]{0.3\textwidth}
        \centering
        \includegraphics[width=\textwidth]{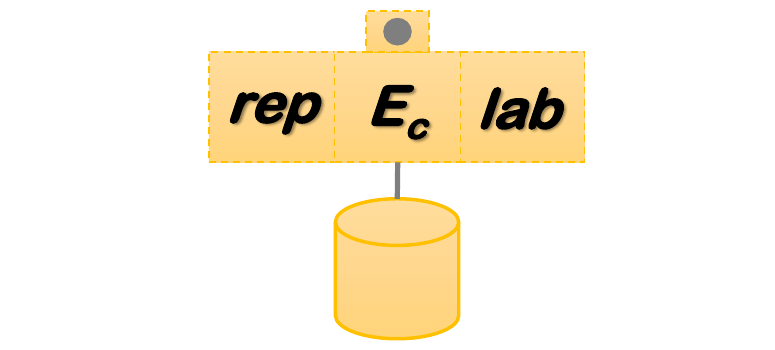}
        \caption{Leaf profile node}
        \label{fig:Leafprofilenode}
    \end{subfigure}
    \caption{The three types of nodes found in the adaptive BCCF-tree}
    \label{fig:TypeNode}
\end{figure}

\begin{itemize}
    \item\textbf{Internal node level} : Nodes at this level contain two pivots $(p_1, p_2)$ and two pointers $(N_L, N_R)$ pointing to the left and right subtrees, as shown in Figure \ref{fig:Internalnode}. The role of this type of node is to connect nodes with different labels or profiles.

    \begin{center}
        \begin{equation*}
            (p_1, p_2, N_L, N_R) \in \mathcal{O}^2 \times \mathcal{N}^2
        \end{equation*}
    \end{center}

    \item \textbf{Internal profile node level}: The nodes at this level are internal nodes, with additional information (see Figure \ref{fig:Internalprofilenode}). These information are: (1) a label $(lab)$ which indicates the root of a subtree of the same profile and (2) a representative feature vector $(rep_f)$. The role of this type of node is to connect nodes with same labels or profiles.

          \begin{center}
              \begin{equation*}
                  (p_1, p_2, lab, rep_f, N_L, N_R) \in \mathcal{O}^2 \times \mathbb{N} \times \mathcal{O} \times \mathcal{N}^2
              \end{equation*}
          \end{center}

    \item \textbf{Leaf profile node level}: Nodes at this level contain a representative feature vector $(rep_f)$, a profile label $(lab)$, and a container $(E_c)$ that consists of a subset of objects whose capacity does not exceed a fixed capacity noted $c_{max}$ (see Figure \ref{fig:Leafprofilenode}). The objective of this node is to avoid over-partitioning of metadata and to keep as much as possible similar metadata in the same container to reduce the search space.

          \begin{center}
              \begin{equation*}
                  (rep_f, lab, E_c) \in \mathcal{O} \times \mathbb{N} \times E, \text{ where, } E_c \leq c_{max}.
              \end{equation*}
          \end{center}

\end{itemize}

\subsection{Label search in Adaptive BCCF-tree}
\label{sec:Labsearch}
In this section, we will present in detail how the label search (see Subsection \ref{step2} and Figure \ref{fig:diagram}) of detected objects was performed on the Adaptive BCCF-tree structure as a solution to the sequential search in the conventional system.

\begin{figure}
    \centering
    \includegraphics[width=0.6\textwidth]{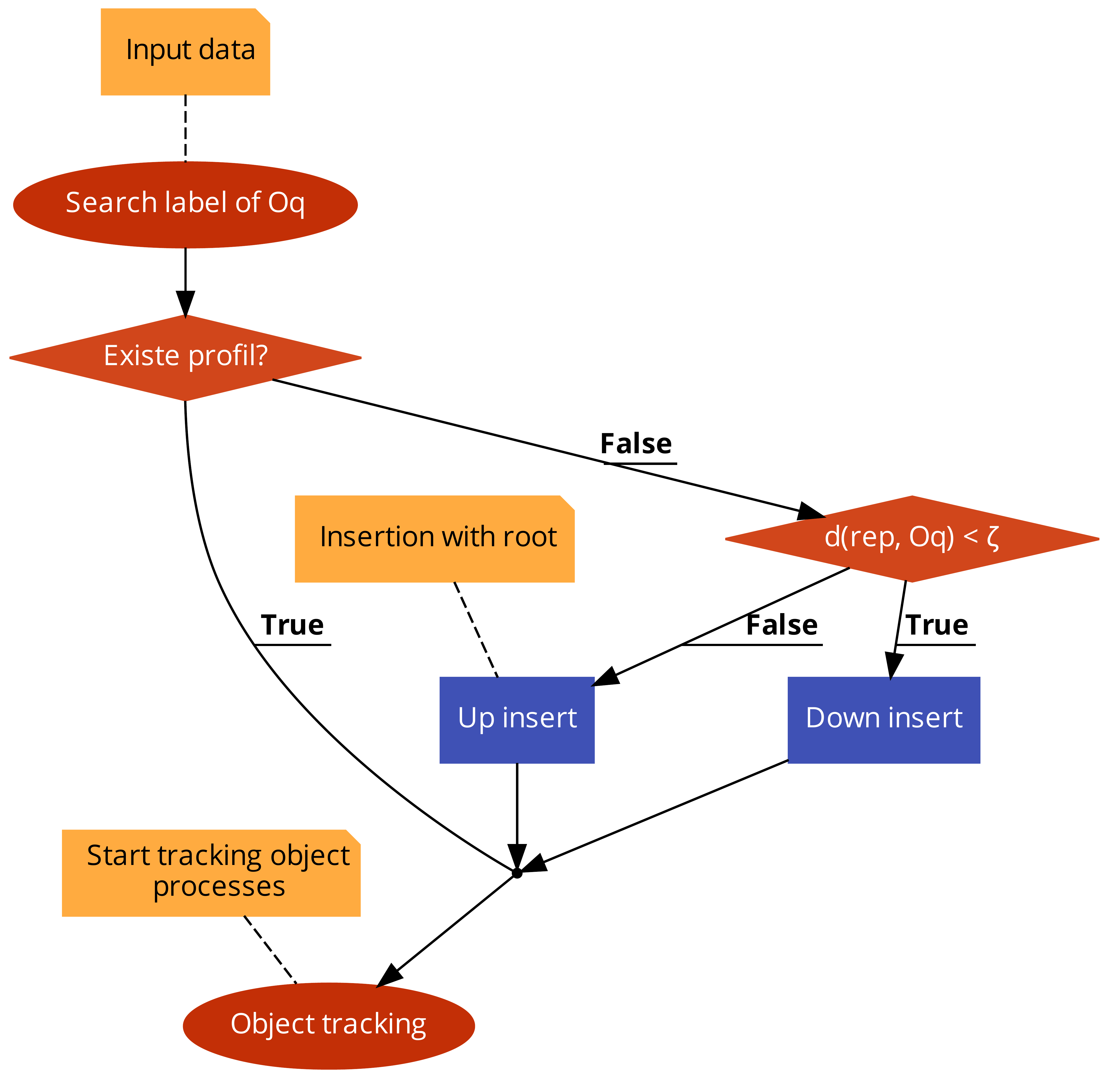}
    \caption{Label search flowchart in the Adaptive BCCF-tree}
    \label{fig:SearchID_FlowChart}
\end{figure}

The main objective of this step is to find the label or identifier of the detected object if it is already tracked previously, to keep the same identifier as much as possible. If the object has appeared for the first time, in this case, this step ensures the determination of a unique identifier. For this purpose, an exact or 1-$nn$ search is triggered on the Adapted BCCF-tree. The flow diagram of this process is shown in Figure \ref{fig:SearchID_FlowChart}, and its algorithmic description is presented in Algorithm \ref{algo:id-search}.

\begin{algorithm}[htp]
    \caption{\texttt{Label search} in the Adaptive BCCF-tree}
    \label{algo:id-search}
    \begin{algorithmic}[1]
        \Require ${\left(
                    \begin{array}{l}
                            N \in \mathcal{N},                          \\
                            o_q \in \mathcal{O},                        \\
                            d : \Objets \times \Objets \to \mathbb{R}^+ \\
                        \end{array}
                    \right)}$
        \Ensure $\left(
            \begin{array}{l}
                    objet-lab \in \mathbb{N}, \\
                    N \in \mathcal{N}
                \end{array}
            \right)$
        \If{N $ = ~\perp $}
        \State $objet-lab \gets \texttt{New-lab}()$

        \State N $= \texttt{New-node-leaf}(o_q, objet-lab, ~\perp)$ 

        \Else
        \While{$N \neq 0$}
        \State $d_1 \gets  d(N.p_1, o_q)$
        \State $d_2 \gets  d(N.p_2, o_q)$
        \If{$d_1  < d_2$}
        \State \textbf{return} \texttt{Label search}($N.N_L, o_q$)
        \Else
        \State \textbf{return} \texttt{Label search}($N.N_R, o_q$)
        \EndIf
        \EndWhile
        \State $d_3 \gets  d(N.rep_f, o_q)$
        \If{$d_3 < \beta$}
        \State $objet-lab \gets  N.lab$
        \Else
        \State $objet-lab \gets \texttt{New-lab}()$
        \If{$d_3 < \zeta$}
        \State \texttt{Down-Insert}($N, o_q, objet-lab$) 
        \Else
        \State \texttt{Up-Insert}($N, o_q, objet-lab$) 
        \EndIf
        \EndIf
        \EndIf
        \State \textbf{return} ($objet-lab, N$)
    \end{algorithmic}
\end{algorithm}

According to Algorithm \ref{algo:id-search}, the search process is done by calculating the distance $d(o_q ,p_i)$ between the query point $o_q$ (the first feature vector of the detected object) and the two current node pivots $(p_1, p_2)$, while going down the tree until reaching the appropriate leaf node that contains the metadata of a most similar identified object. At the leaf level, a second distance $d(o_q, rep_i)$ will be calculated between the query point $o_q$ and the leaf node representative $rep_f$ that was reached. The computed distance will be compared with a previously defined similarity threshold $\beta$ to determine whether this object belongs to the same reached profile or not.

If the distance is less than the threshold $d(o_q, rep_i) < \beta$, it means that the detected object belongs to the same leaf node profile reached. In this case, the identifier of reached leaf node will be assigned to the detected object to start its tracking processes. If the distance is greater than the threshold $d(o_q, rep_i) \ge \beta$, we determine that the detected object is not the same object of the reached profile, but it is similar. In this case, a new unique identifier has been generated for this object. After that, we try to reserve a profile leaf node for this object in the right place according to a second similarity threshold $T$. The purpose of this second threshold is to determine whether this object should be kept close to the reached profile leaf node or not according to the two insertion strategies: insertion down or insertion up presented in Algorithms \ref{alg:InsertDown} and \ref{alg:InsertUP}, respectively.\\

\begin{enumerate}
    \item \textit{Down insertion strategy}
\end{enumerate}

In the case of a strong similarity between the new object and the profile reached during the search ($d(N.rep_f, o_q) < \zeta $), the insertion down process on this object is triggered. This insertion strategy consists in grouping the new detected object with the profile closest to it. The objective is to guarantee a better placement of the indexed profiles, to reduce as much as possible the search space, and therefore the search time. The pseudocode of this strategy is presented in Algorithm \ref{alg:InsertDown}.

\begin{algorithm}[htp]
    \caption{\texttt{Down insertion}}
    \label{alg:InsertDown}
    \begin{algorithmic}[1]
        \Require ${\left(
                    \begin{array}{l}
                            N \in \mathcal{N},       \\
                            o_q \in \mathcal{O},     \\
                            objet-lab \in \mathbb{N} \\
                        \end{array}
                    \right)}$
        \Ensure $\left(
            \begin{array}{l}
                    N \in \mathcal{N}, \\
                    node_{(oq)} \in \mathcal{N}
                \end{array}
            \right)$

        \State $node_{(o_q)} = \texttt{New-node-profile-leaf}(o_q, objet-lab,~\perp)$

        \State $intenode = \texttt{New-node-internal}(o_q, N.rep_f, node_{(o_q)}, N)$

        \State $\texttt{Pointerswitch}(N, intenode)$

        \State \textbf{return} ($N, node_{(o_q)}$)

    \end{algorithmic}
\end{algorithm}

The principle of this strategy is to create a subtree rooted at a new internal node ($intenode$) pointing respectively to the internal profile node of the most similar object ($N$) to the inserted object and a new profile leaf node ($node_{(o_q)}$) that corresponds to the inserted object. The two pivots of the root of this subtree ($intenode$) are respectively the representatives of new object ($o_q$) and the most similar profile ($N.rep_f$).\\

\begin{enumerate}[resume]
    \item \textit{UP insertion strategy}
\end{enumerate}

In the case of a strong dis-similarity between the new object and the profile reached during the search ($d(N.rep_f, o_q) \geq \zeta$), a grouping of the latter with the root is applied (insertion at the top), which makes it possible to create distinct groups of profiles that are not similar according to Algorithm \ref{alg:InsertUP}.


As shown in Algorithm \ref{alg:InsertUP}, a distance $d(o_q, p_i)$ between the inserted object and the two pivots (p$_1$, p$_2$) of the root will be calculated. If the distance $d(o_q, p_1)$ is less than or equal to the distance $d(o_q, p_2)$, a new root will be created. The two pivots of the new root (p$_1$, p$_2$) are respectively the representative of the new profile ($o_q$) and the left pivot (p$_1$) of the old root. Concerning the pointers of the new root ($N_L$, $N_R$), the left pointer ($N_L$) is pointed to a new profile leaf node that corresponds to the newly inserted object ($node_{(oq)}$), while the right pointer ($N_G$) is pointed to the old root ($N$). Otherwise, i.e., the distance $d(rep_i, p_1)$ is greater than the distance $d(rep_i, p_2)$. In this case, also a new root will be created such that the two pivots (p$_1$, p$_2$) of the new root are the representative of the new profile ($o_q$) and the right pivot (p$_2$) of the old root. Regarding the pointers of the new root ($N_L$, $N_R$), it is the reverse of the previous case where the left pointer ($N_L$) is pointed to the old root ($N$), while the right pointer (D$_G$), is pointed to a new profile leaf node that corresponds to the newly inserted object ($node_{(oq)}$).

\begin{algorithm}[htp]
    \caption{\texttt{UP insertion}}
    \label{alg:InsertUP}
    \begin{algorithmic}[1]
        \Require ${\left(
                    \begin{array}{l}
                            N \in \mathcal{N},       \\
                            o_q \in \mathcal{O},     \\
                            objet-lab \in \mathbb{N} \\
                        \end{array}
                    \right)}$
        \Ensure $\left(
            \begin{array}{l}
                    N \in \mathcal{N},          \\
                    node_{(oq)} \in \mathcal{N} \\
                \end{array}
            \right)$

        \State $d_1 \gets  d(N.p_1, o_q)$
        \State $d_2 \gets  d(N.p_2, o_q)$
        \If{d$_1 < $ d$_2$}
        \State      $node_{(o_q)} = \texttt{New-node-profile-leaf}(o_q, objet-lab,~\perp)$

        \State    root = $\texttt{New-node-internal}(N.p_2, o_q, N, node_{(o_q)})$ 

        \Else
        \State     $node_{(o_q)} = \texttt{New-node-profile-leaf}(o_q, objet-lab,~\perp)$

        \State     root = $\texttt{New-node-internal}(N.p_2, o_q, N, node_{(o_q)})$ 

        \EndIf

    \end{algorithmic}
\end{algorithm}

\subsection{Labeled metadata insertion in Adaptive BCCF-tree}

During the tracking process (see Subsection \ref{step3} and Figure \ref{fig:diagram}), the extracted metadata on the tracked object is collected in a common set. After the end of the tracking process, this metadata set is transmitted to the automatic labeling system for indexing it. For this purpose, the automatic labeling system starts the insertion process described in Algorithm \ref{algo:partitionnment}.

\begin{algorithm}[htp]
    \caption{\texttt{Insertion by batch}}
    \label{algo:partitionnment}
    \begin{algorithmic}[1]
        \Require ${\left(
                    \begin{array}{l}
                            N \in \mathcal{N},     \\
                            bag \in   \mathcal{O}, \\
                            objet-lab \in \mathbb{N}
                        \end{array}
                    \right)}$
        \Ensure $\left(
            \begin{array}{l}
                    N \in \mathcal{N} \\
                \end{array}
            \right)$

        \If{$\|N.E_c ~\cap~ bag\| \geq C_{max}$}
        \State $(Cluster_1, Cluster_2, C_1, C_2) = \texttt{Partition}(N.E_c ~\cap bag)$
        \State $N.N_L \gets \texttt{New-node-leaf} (C1, N.lab, Cluster_1)$

        \State $N.N_R \gets \texttt{New-node-leaf} (C2, N.lab, Cluster_2)$


        \State $rep_f \gets \texttt{mean}(N.p_1, N.p_2)$

        \State $N \gets \texttt{ChangeToInternalprofile}(C_1, C_2, rep_f, N.N_L, N.N_R )$

        \Else

        \State $N.E_c \gets  N.E_c ~\cap~ bag$

        \EndIf

    \end{algorithmic}
\end{algorithm}

According to Algorithm \ref{algo:partitionnment}, if the current size of the corresponding leaf node container together with the size of the new data collection (bag) are not greater than the cardinal limit of the  container $c_{max}$, the new metadata collection is inserted directly into the same leaf node container. Otherwise, the partitioning process is triggered on these data (i.e., the received metadata and the corresponding profile leaf node metadata). After the convergence of the latter, two clusters and two centers are obtained. Next, the partitioned leaf node becomes an internal profile node, and the two resulting centers represent its pivots. Finally, two leaf profile nodes are created, where the two clusters become their containers. These leaf profile nodes are rooted at the partitioned node.


\section{EXPERIMENTS AND COMPARISON}
\label{section5}

The proposed system is implemented in Python on a Linux workstation with an Intel{\textregistered} Core$^{TM}$ i5, and 4 GB RAM memory capacity. 

In order to evaluate the performance of our system, we carried out all experiments on real data with different distributions using the MOT16 dataset \cite{milan2016mot16}. This database contains 14 video sequences (7 training, 7 testing) in unconstrained environments, filmed with static and mobile cameras. All sequences have been annotated with great precision, strictly following a well-defined protocol \cite{milan2016mot16}.

Table \ref{tab:dataset} shows the characteristics of these datasets, the similarity thresholds used, and the maximum capacity of the centenaries defined in the experiment.

    {
        \begin{table}
            \caption{Dataset characteristics}
            \label{tab:dataset}
            \centering
            \begin{tabular}{lllllll|}
                \hline
                $Dataset$  & $\#Objects \times 100$ & $Dimension$ & $\beta$ & $T$ & $C_{max}$ \\ \hline
                MOT16$-06$ & 7 963 000              & 128         & 0.2     & 0.6 & 892       \\

                MOT16$-07$ & 1 040 400              & 128         & 0.3     & 0.8 & 1 020     \\
                MOT16$-12$ & 488 200                & 128         & 0.2     & 0.8 & 699       \\
                MOT16$-14$ & 1 150 000              & 128         & 0.3     & 0.8 & 1 072     \\ \hline
            \end{tabular}
        \end{table}
    }

The evaluation of our proposed solution involves two steps:

\begin{description}
    \item[Step 1:]~Evaluate the Adaptive BCCF-tree in terms of construction cost and search efficiency.
    \item[Step 2:]~Evaluate the tracking system's performance after introducing the Adaptive BCCF-tree and compare it with the system without indexing and other recent systems.
\end{description}

\begin{figure*}[h]
    \centering
    \begin{subfigure}[b]{0.49\textwidth}
        \centering
        \includegraphics[width=\textwidth]{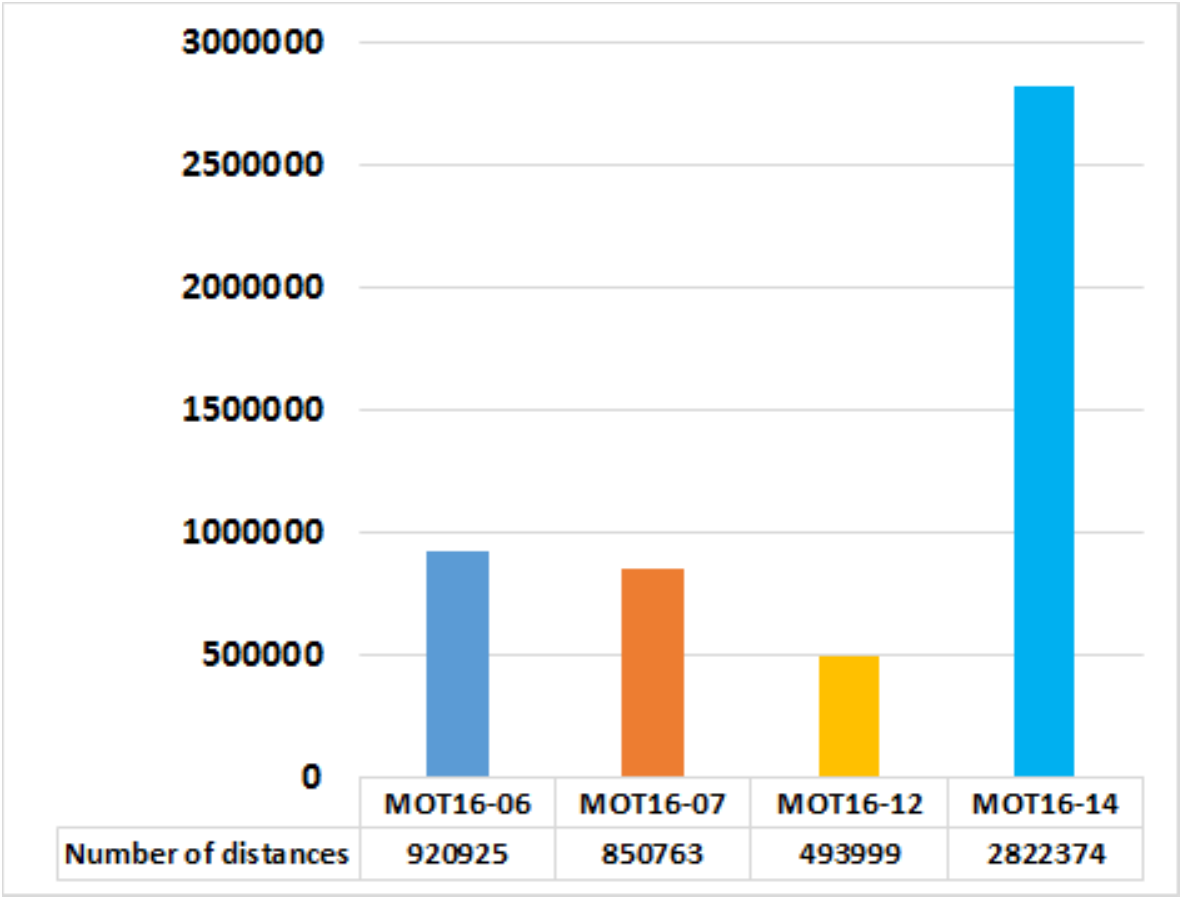}
        \caption{Number of distances}
        \label{fig:dist1}
    \end{subfigure}
    \hfill
    \begin{subfigure}[b]{0.49\textwidth}
        \centering
        \includegraphics[width=\textwidth]{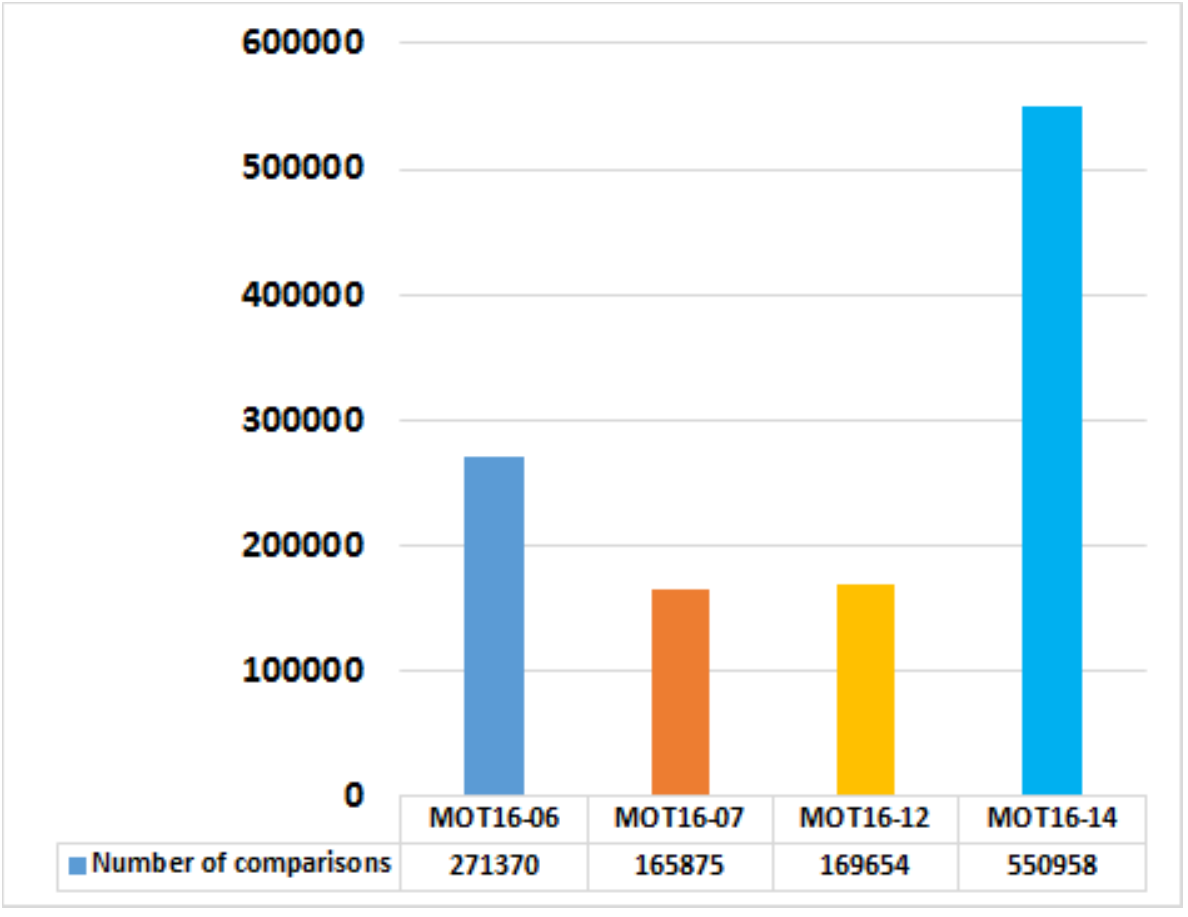}
        \caption{Number of comparisons}
        \label{fig:comp}
    \end{subfigure}
    \hfill
    \caption{Number of distances computed and comparisons performed for the construction of the Adaptive BCCF-tree}
    \label{fig:dist_comp}
\end{figure*}

\begin{figure*}
    \centering
    \begin{subfigure}[b]{0.49\textwidth}
        \centering
        \includegraphics[width=\textwidth]{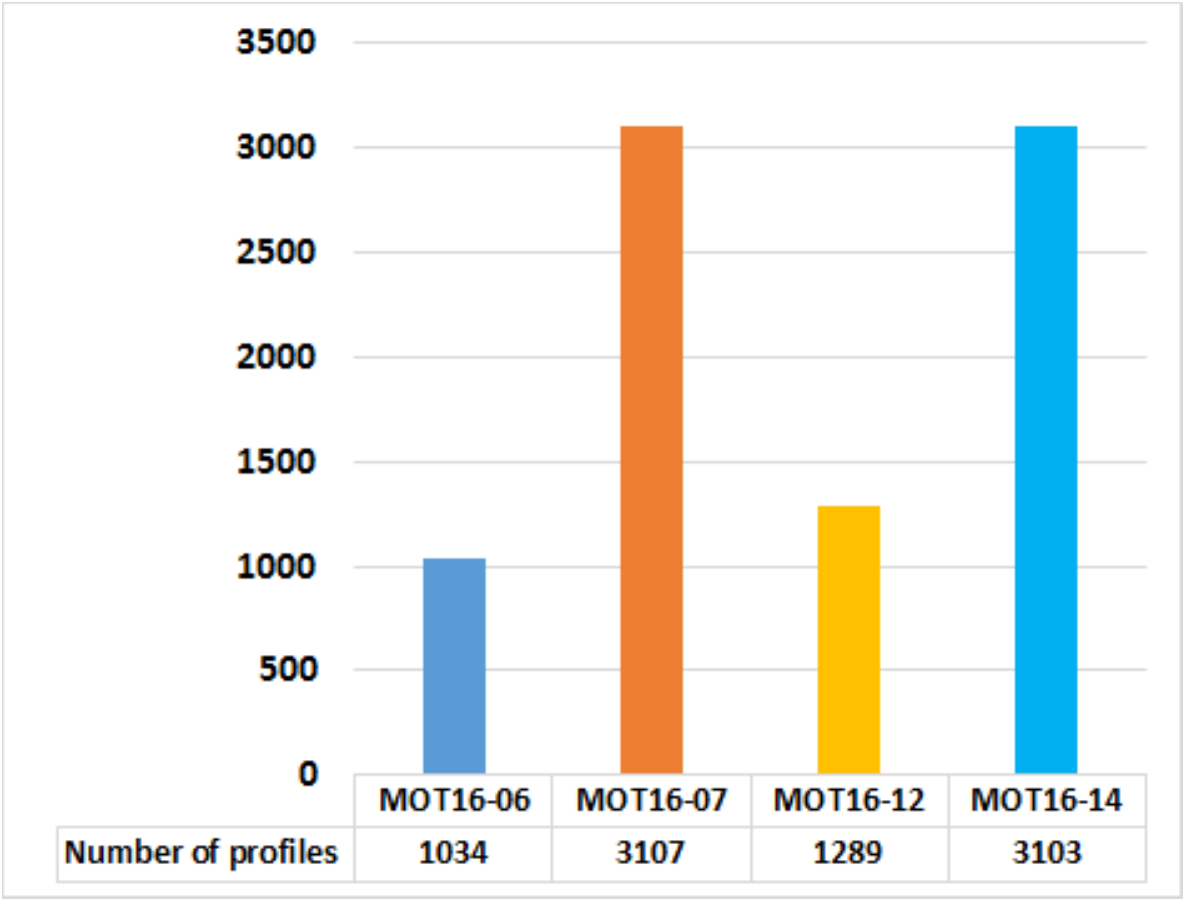}
        \caption{Number of profiles}
        \label{fig:nombre profil}
    \end{subfigure}
    \hfill
    \begin{subfigure}[b]{0.49\textwidth}
        \centering
        \includegraphics[width=\textwidth]{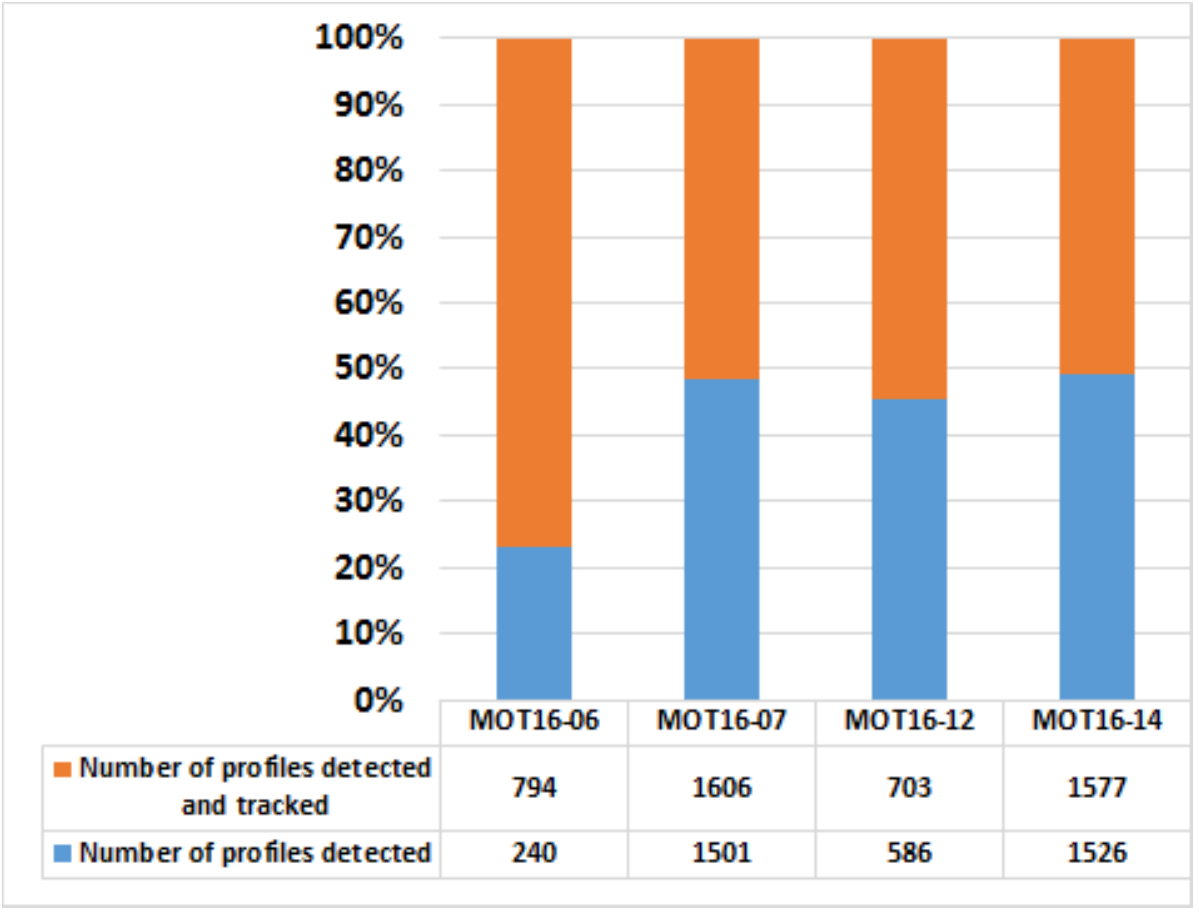}
        \caption{Number of profiles detected and tracked}
        \label{fig:detecte suivi }
    \end{subfigure}
    \hfill
    \caption{Evaluation of the number of profiles for each tree created}
    \label{fig:nombre_profil}
\end{figure*}

\begin{figure*}
    \centering
    \begin{subfigure}[b]{0.49\textwidth}
        \centering
        \includegraphics[width=\textwidth]{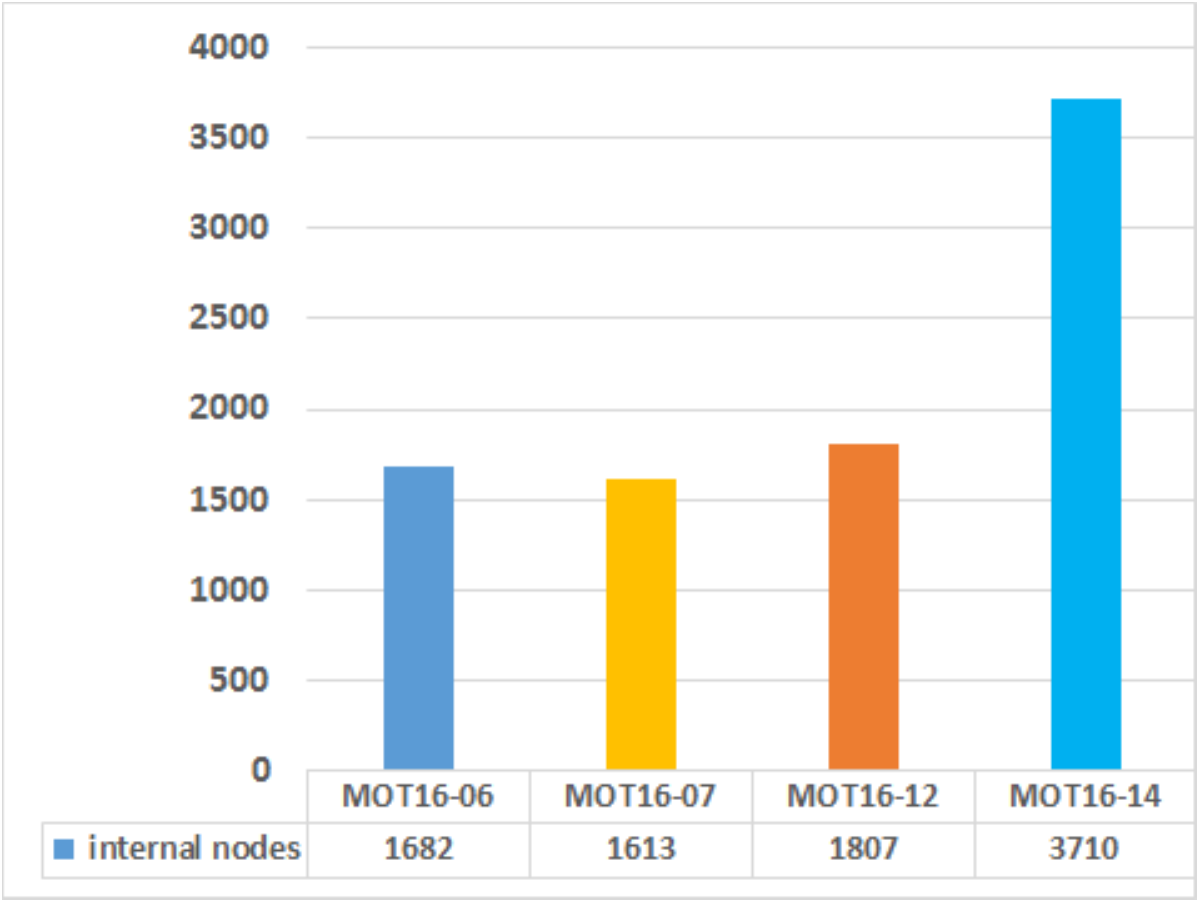}
        \caption{Number of internal nodes}
        \label{fig:noued interne}
    \end{subfigure}
    \hfill
    \begin{subfigure}[b]{0.49\textwidth}
        \centering
        \includegraphics[width=\textwidth]{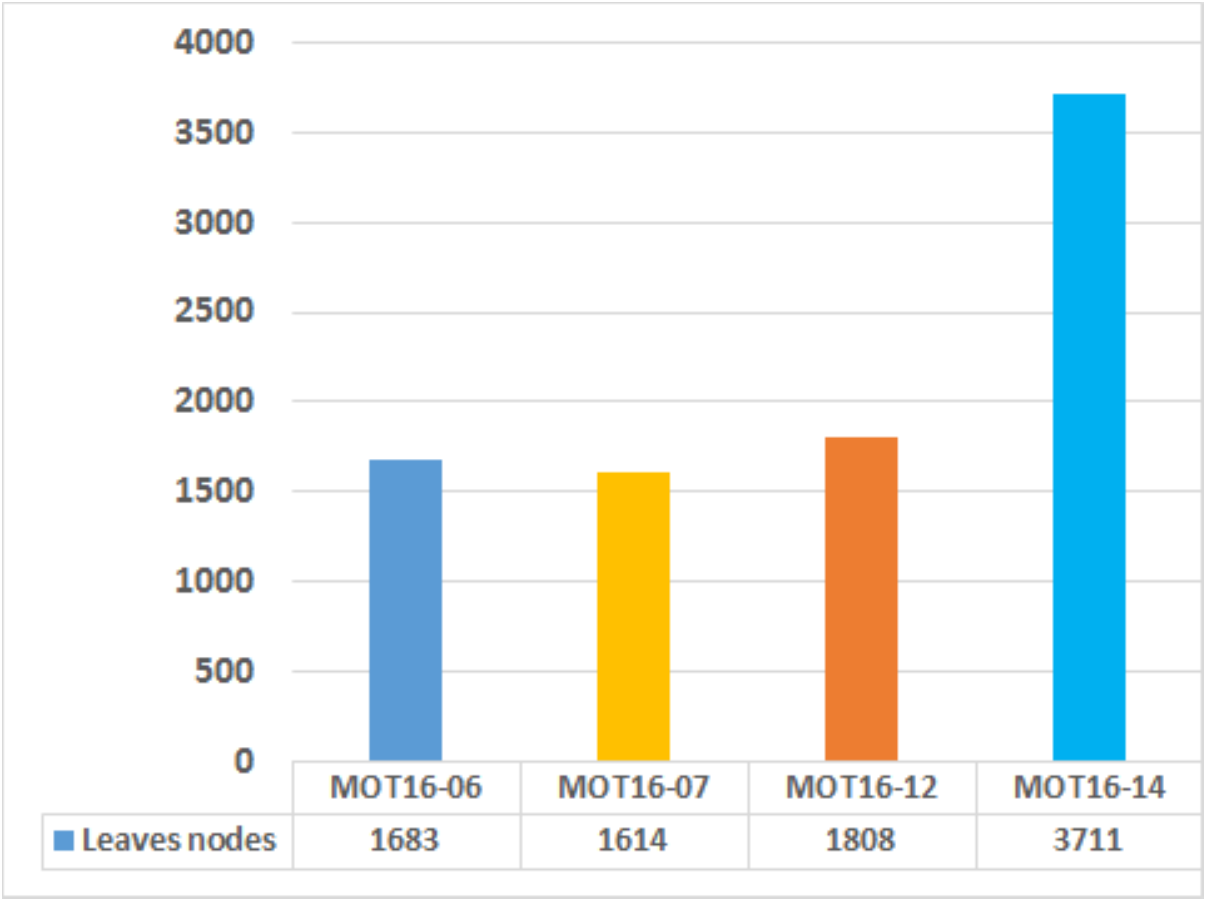}
        \caption{Number of leaf nodes}
        \label{fig:noued feuille}
    \end{subfigure}
    \hfill
    \begin{subfigure}[b]{0.49\textwidth}
        \centering
        \includegraphics[width=\textwidth]{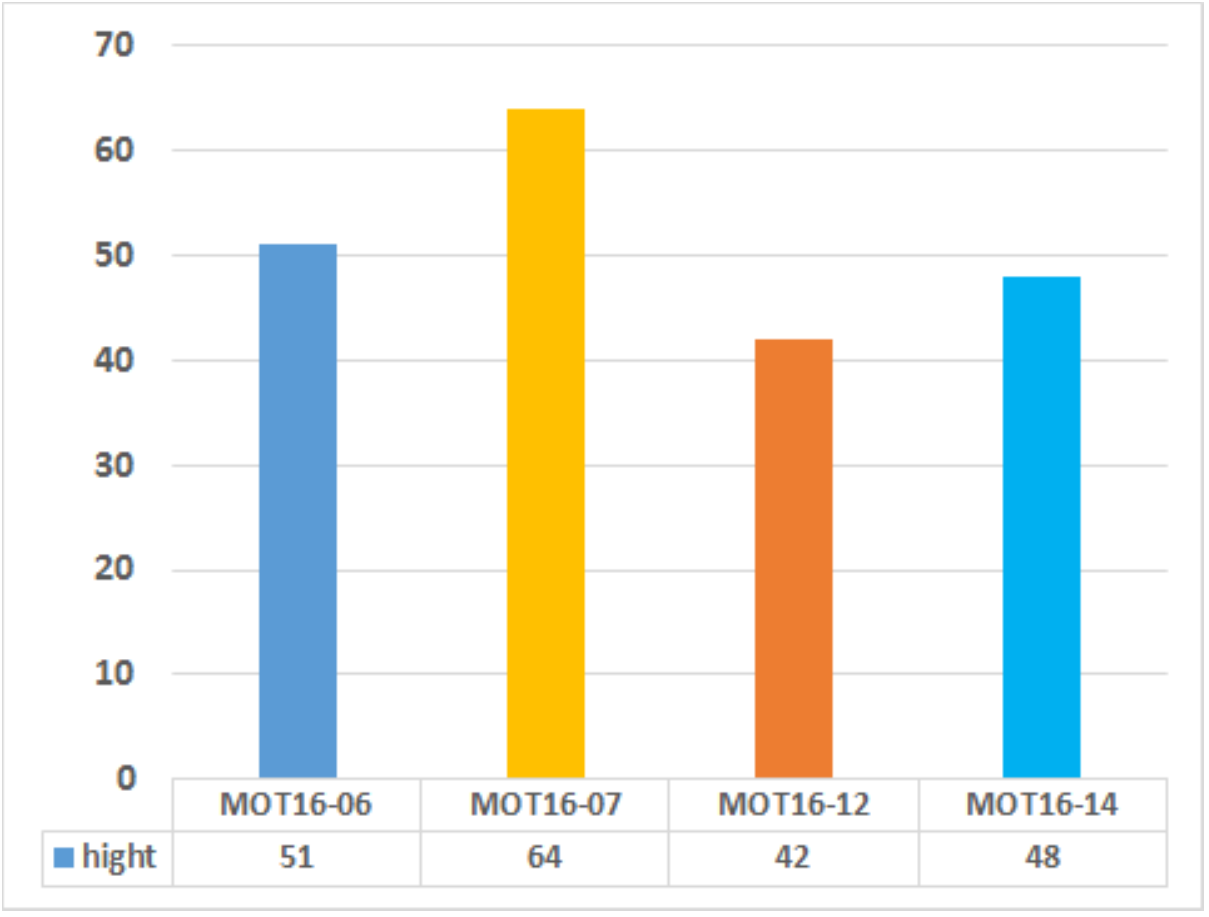}
        \caption{Height of the tree}
        \label{fig:hauteur}
    \end{subfigure}
    \caption{Evaluation of the index structure}
    \label{fig:eval_indice}
\end{figure*}

\subsection{Adaptive BCCF-tree evaluation}
\label{sec:index-eva}

This section presents the results obtained by the Adaptive BCCF-tree structure on the MOT16 dataset test sequences $06$, $07$, $12$, and $14$. To perform this evaluation, we need a large dataset, which was not provided by this dataset or other public dataset. For this purpose, we performed the evaluation on the previously mentioned sequences multiplied $\times 100$ times for each sequence.

The number of distances and comparisons made during the construction of the Adaptive BCCF-tree are computed in order to evaluate its construction cost. The obtained results are shown in Figure \ref{fig:dist_comp} where the size of the containers is fixed by $c_{max}= \sqrt{n}$ ($n = \|E\|$, the size of the data set) according to \cite{benrazek2020efficient}.

Knowing that the construction cost depends on comparing the distances between the inserted objects and the pivots of each internal node, the results show that the construction cost increases with the increase of the number of indexed objects, as we observe in the sets MOT16$-12$ and MOT16$-14$ in Figure \ref{fig:dist_comp}. On the other hand, the construction cost also increases with the number of tracked objects. Figure \ref{fig:nombre_profil} present the number of detected objects, and the number of tracked objects in each dataset. According to Figure \ref{fig:dist_comp}, \ref{fig:nombre_profil} and Table \ref{tab:dataset}, we observe that the number of objects in the set MOT16$-06$ is greater than in the set MOT16$-07$ (see Table \ref{tab:dataset}), while the number of tracked objects in the MOT16$-07$ set is greater than that in the MOT16$-06$ set (see Figure \ref{fig:nombre_profil}) and the results show that the cost of constructing the index in the MOT16$-06$ set is more expensive than in the MOT16$-07$ set (see Figure \ref{fig:dist_comp}). This increase is due to the need to partition the data as shown in Figure \ref{fig:eval_indice}, where the height, number of leaf nodes and internal nodes of the tree in the MOT16$-06$ set are almost equal or closer to those of the tree in the MOT16$-07$ set, which has a large number of objects.

Now, we pass to the evaluation of the Adaptive BCCF-tree in terms of search efficiency. This evaluation is conducted in two parts:

\begin{itemize}
    \item The first part is to evaluate the search time during the operation of the tracking system, in other words, the online or real-time search. The results of this study are presented in Section \ref{subsec:onlineSearch}.

    \item The second part is to evaluate the search time at the end of the system operation, i.e., the offline search. The experimental analysis of this study is presented in Section \ref{subsec:offSearch}.
\end{itemize}

\subsubsection{Online Research Evaluation}
\label{subsec:onlineSearch}

\begin{figure*}
    \centering
    \begin{subfigure}[b]{0.48\textwidth}
        \centering
        \includegraphics[width=\textwidth]{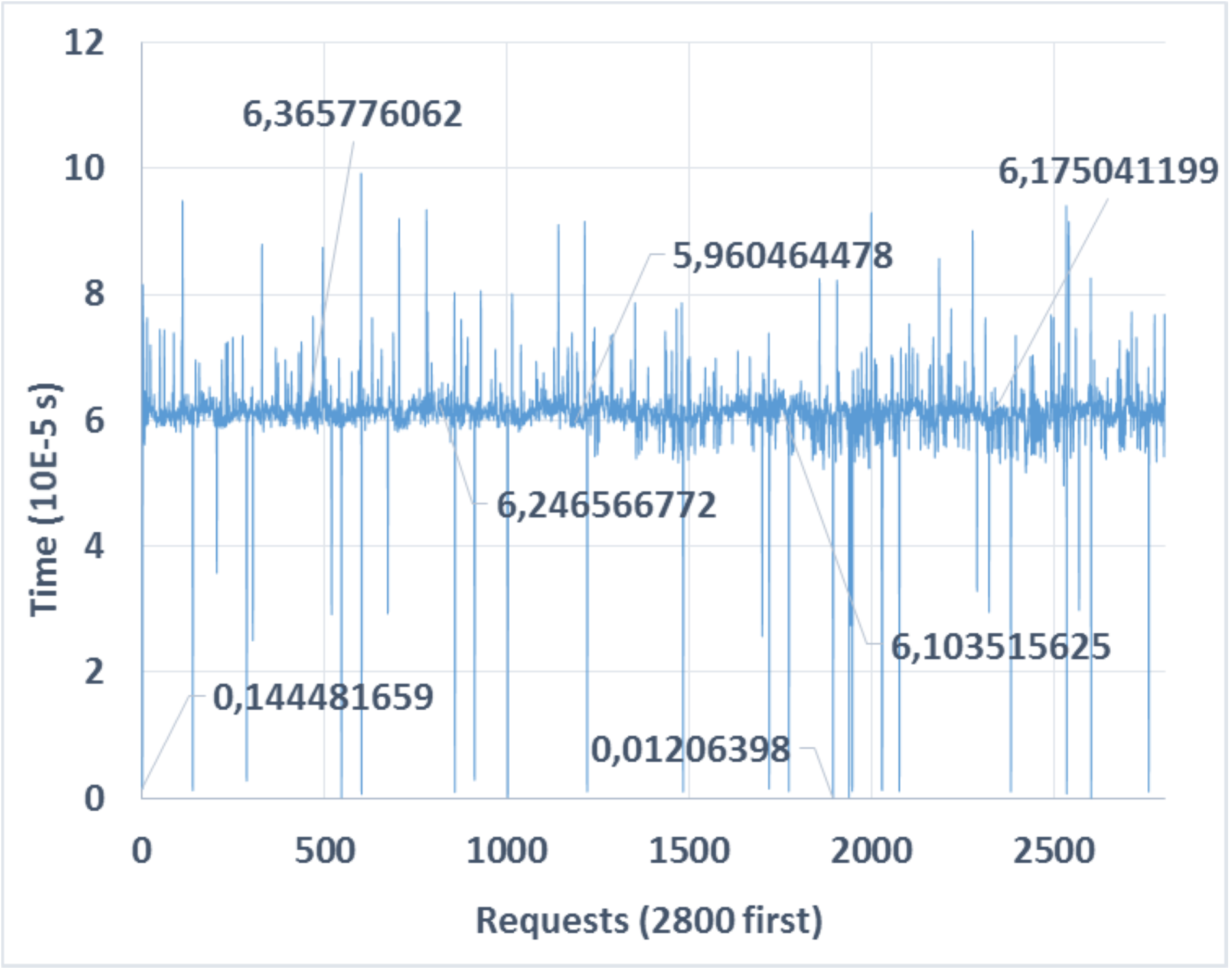}
        \caption{MOT16$-06$}
        \label{fig:search06}
    \end{subfigure}
    \hfill
    \begin{subfigure}[b]{0.48\textwidth}
        \centering
        \includegraphics[width=\textwidth]{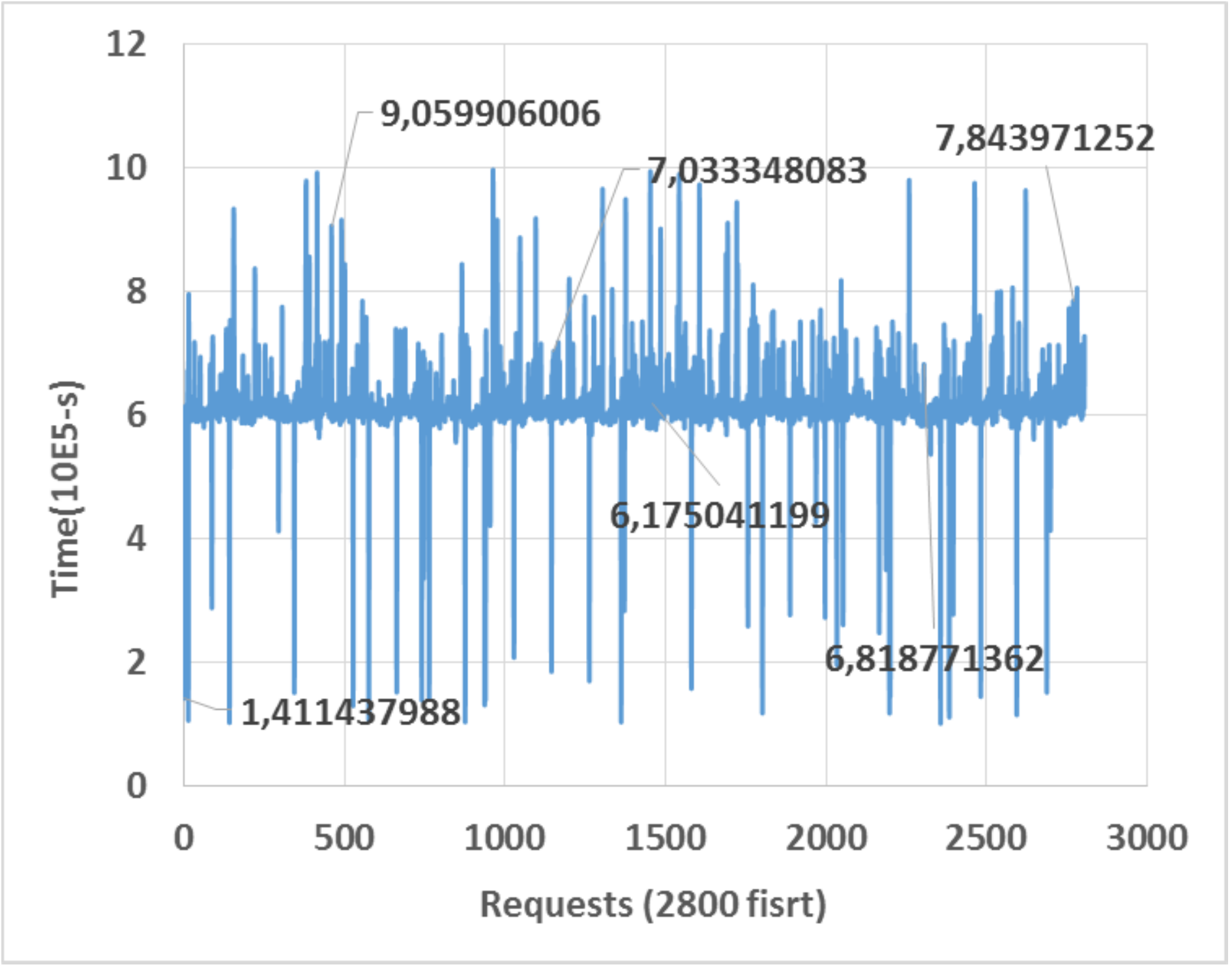}
        \caption{MOT16$-07$}
        \label{fig:search07}
    \end{subfigure}
    \hfill
    \begin{subfigure}[b]{0.48\textwidth}
        \centering
        \includegraphics[width=\textwidth]{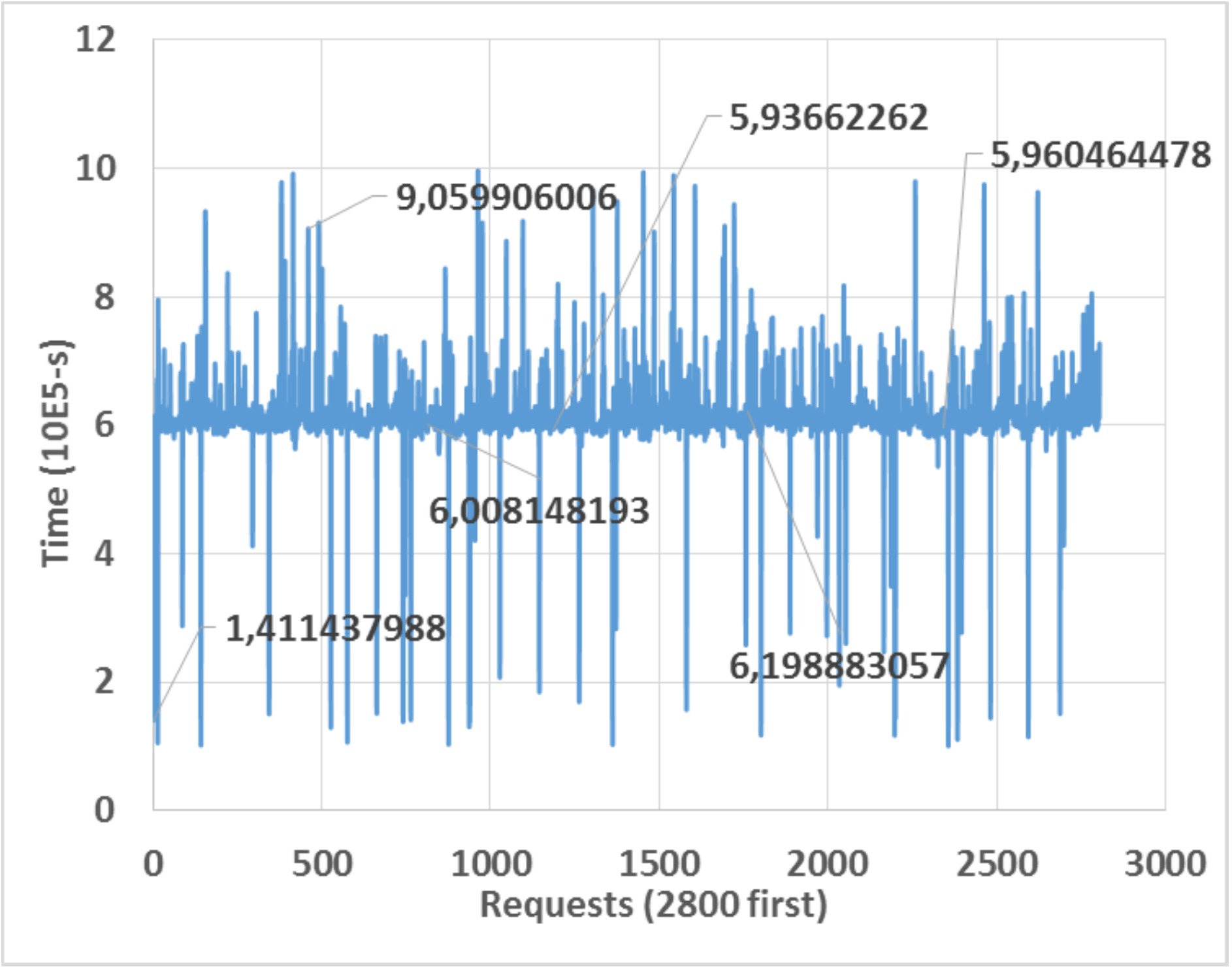}
        \caption{MOT16$-12$}
        \label{fig:search12}
    \end{subfigure}
    \hfill
    \begin{subfigure}[b]{0.48\textwidth}
        \centering
        \includegraphics[width=\textwidth]{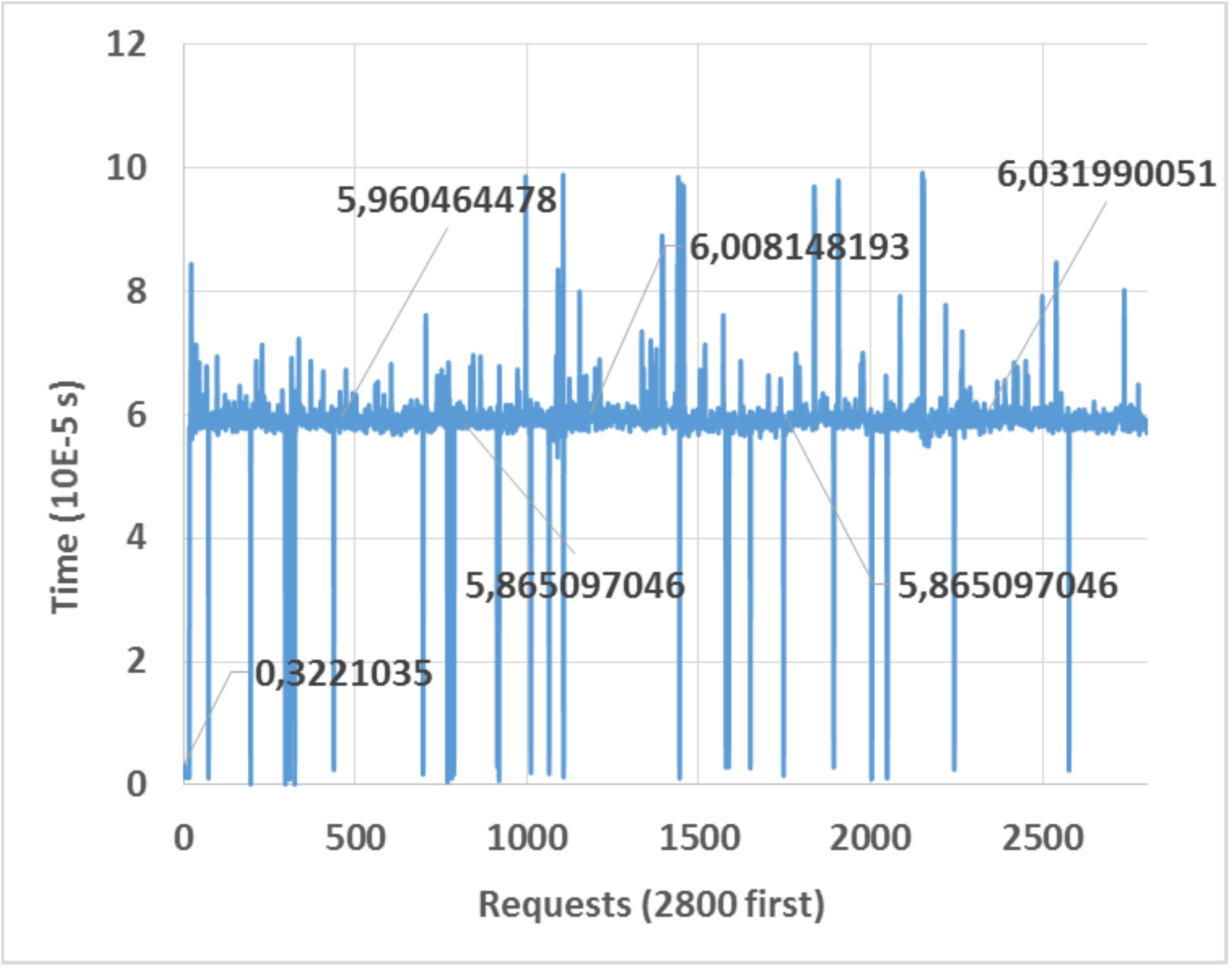}
        \caption{MOT16$-14$}
        \label{fig:search14}
    \end{subfigure}
    \caption{Online search time}
    \label{fig:online_search}
\end{figure*}

\begin{figure*}
    \centering
    \begin{subfigure}[b]{0.48\textwidth}
        \centering
        \includegraphics[width=\textwidth]{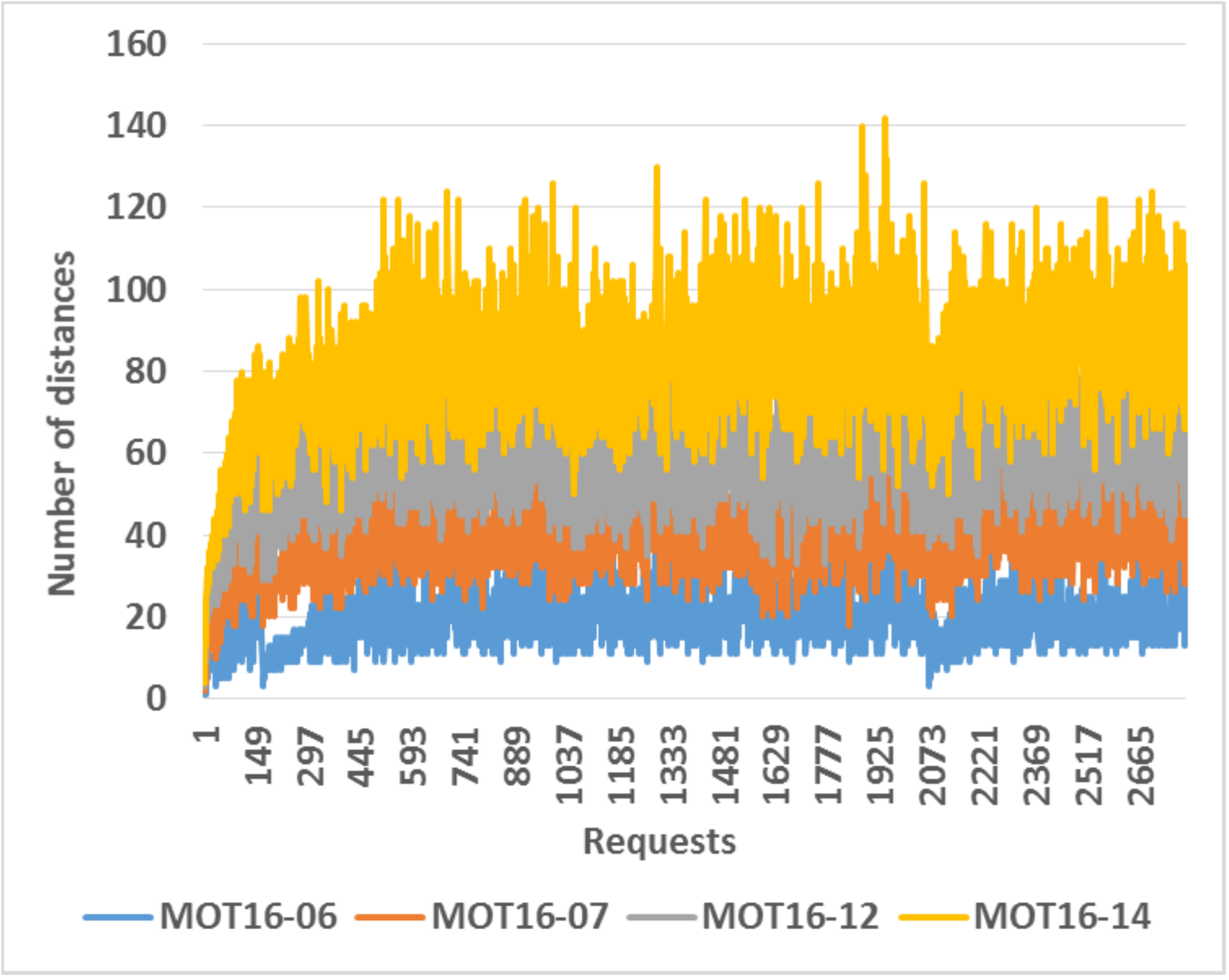}
        \caption{Number of distances}
        \label{fig:dist_online1}
    \end{subfigure}
    \hfill
    \begin{subfigure}[b]{0.48\textwidth}
        \centering
        \includegraphics[width=\textwidth]{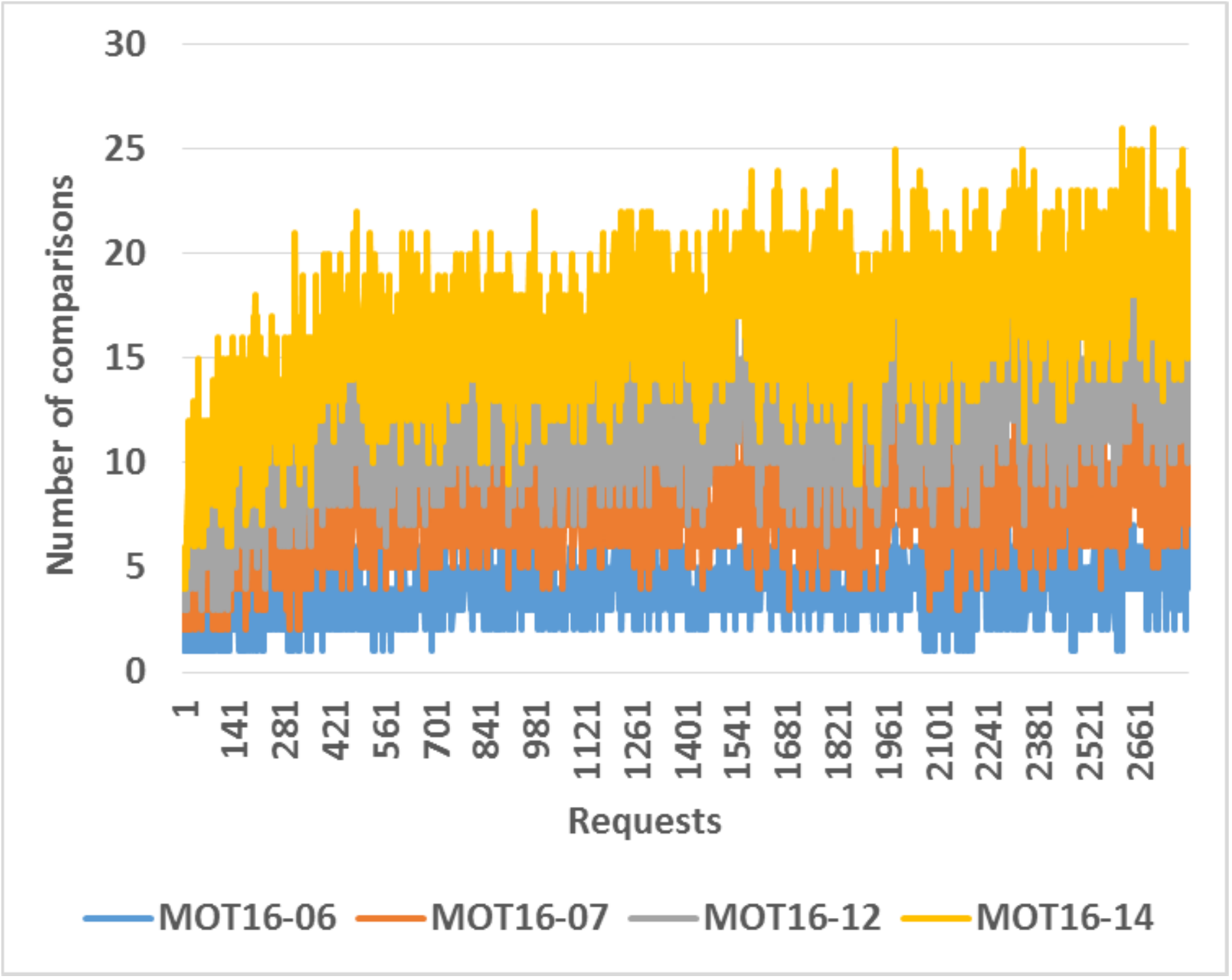}
        \caption{Number of comparisons}
        \label{fig:componline1}
    \end{subfigure}
    \hfill
    \caption{Online search cost}
    \label{fig:dist_comp_on}
\end{figure*}

This section presents the real-time search performance of the proposed structure on different datasets, and the results of the experiments are shown in Figures \ref{fig:online_search} and \ref{fig:dist_comp_on}. For a better interpretation of the results, we summarized them in Table \ref{tab:recherche_online}, which presents the average of the values obtained for each of the following evaluation criteria:

\begin{itemize}
    \item The average number of distances calculated.
    \item The average number of comparisons performed.
    \item The average search time.
\end{itemize}

{
\begin{table}
    \caption{Online average search results}
    \label{tab:recherche_online}
    \centering
    \begin{tabular}{lcccc}
        \hline
                       & MOT16$-06$ & MOT16$-07$ & MOT16$-12$ & MOT16$-14$ \\ \hline
        \#Distances    & 22         & 20         & 9          & 22         \\
        \#Comparaisons & 4          & 4          & 4          & 5          \\
        Time (10E-5s)  & 6.11       & 5.91       & 6.03       & 5.87       \\ \hline
    \end{tabular}

\end{table}
}
From Figure \ref{fig:online_search}, the results show on the one hand that the search time does not exceed 10$\times10^{-5}s$ in the worst case where the search time is between 9.77$\times10^{-5}s$ and 9.98$\times10^{-5}s$. This means that the most similar object to the object in question is at the maximum depth of the index structure. On the other hand, the results present that the search time decreased until reaching the minimum value of 0.01$\times10^{-5}s$. This corresponds to the best case where the searched objects are located closer to the root due to their merging by the up insertion technique presented in Section \ref{sec:Labsearch}. According to Table \ref{tab:recherche_online}, the results show that the average search time varies in the range [5.87$\times10^{-5}s$, 6.11$\times10^{-5}s$]. An increase in search time leads to an increase in the number of distances and comparisons respectively and vice versa, as shown in Figure \ref{fig:dist_comp_on}.

\subsubsection{Offline Research Evaluation}
\label{subsec:offSearch}

\begin{figure*}
    \centering
    \begin{subfigure}[b]{0.49\textwidth}
        \centering
        \includegraphics[width=\textwidth]{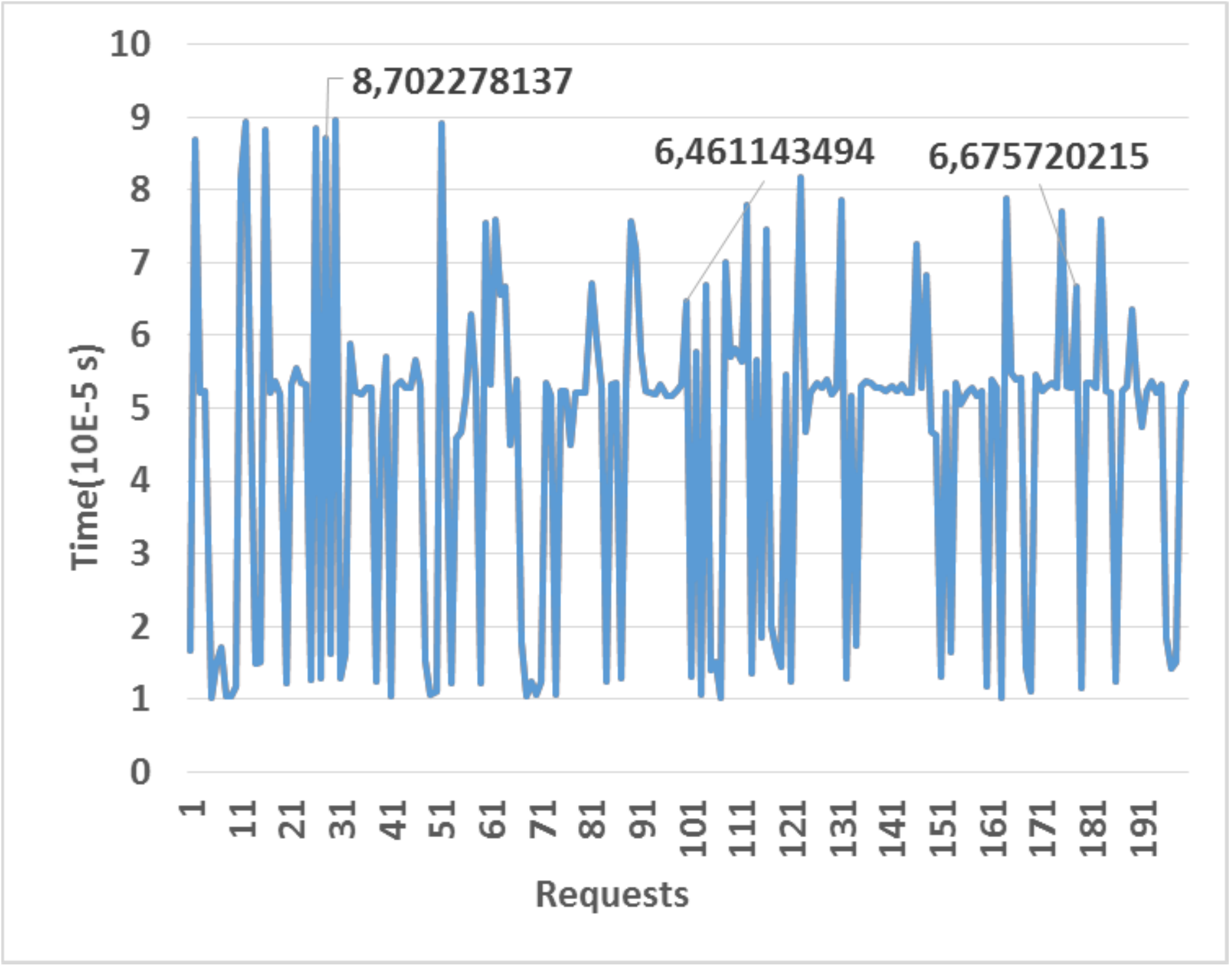}
        \caption{MOT16$-06$}
        \label{fig:off_search06}
    \end{subfigure}
    \hfill
    \begin{subfigure}[b]{0.49\textwidth}
        \centering
        \includegraphics[width=\textwidth]{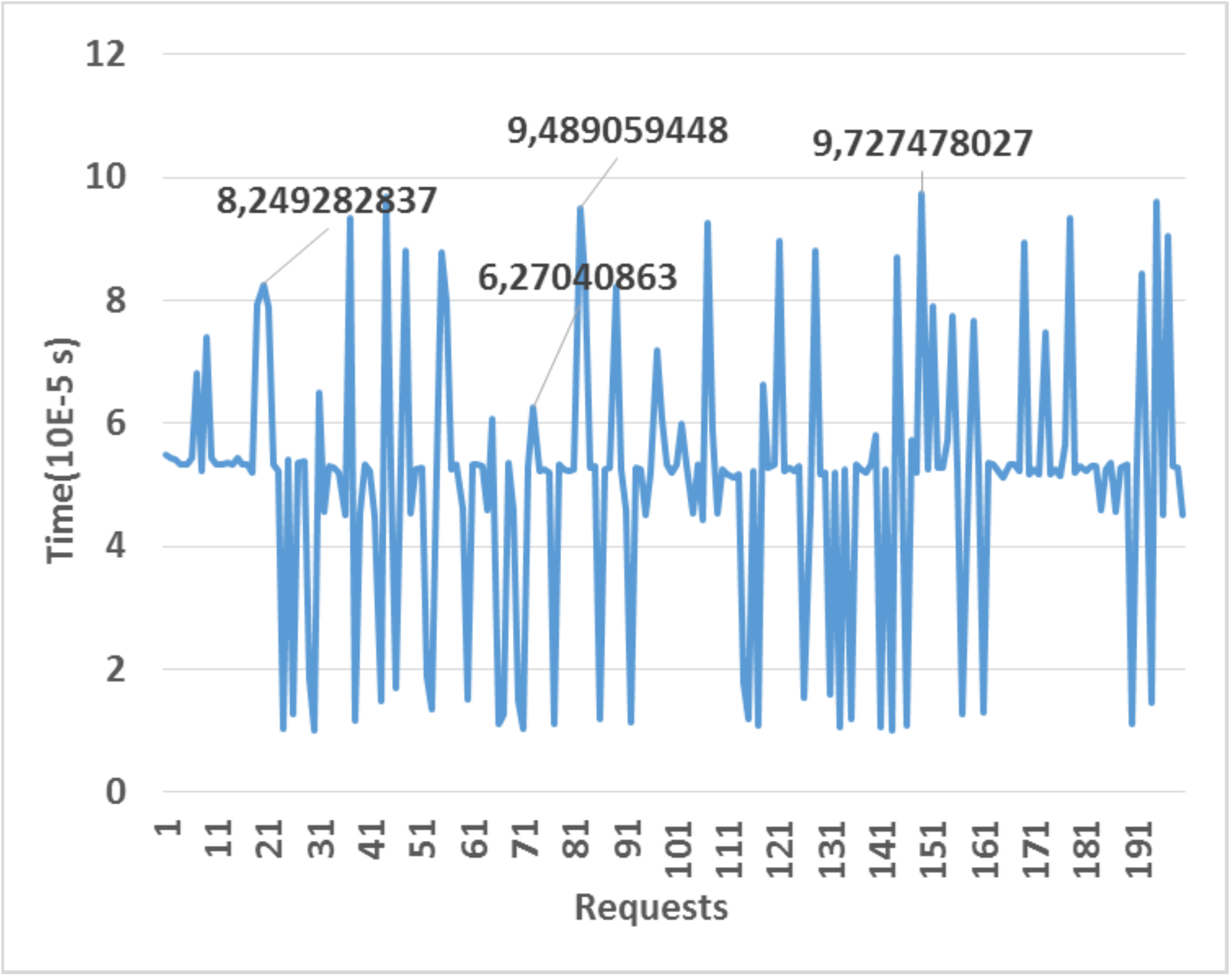}
        \caption{MOT16$-07$}
        \label{fig:off_search07}
    \end{subfigure}
    \hfill
    \begin{subfigure}[b]{0.49\textwidth}
        \centering
        \includegraphics[width=\textwidth]{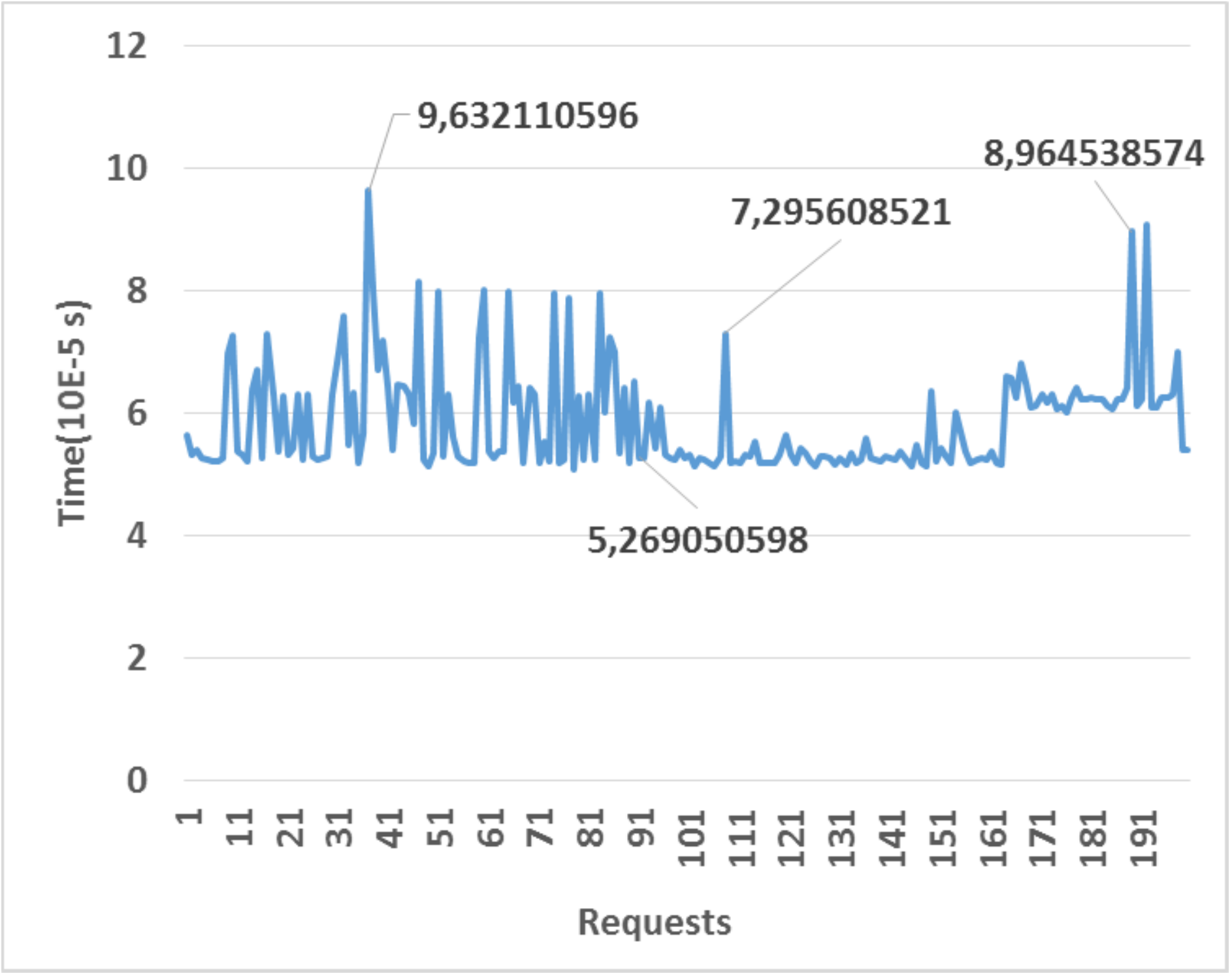}
        \caption{MOT16$-12$}
        \label{fig:off_search12}
    \end{subfigure}
    \hfill
    \begin{subfigure}[b]{0.49\textwidth}
        \centering
        \includegraphics[width=\textwidth]{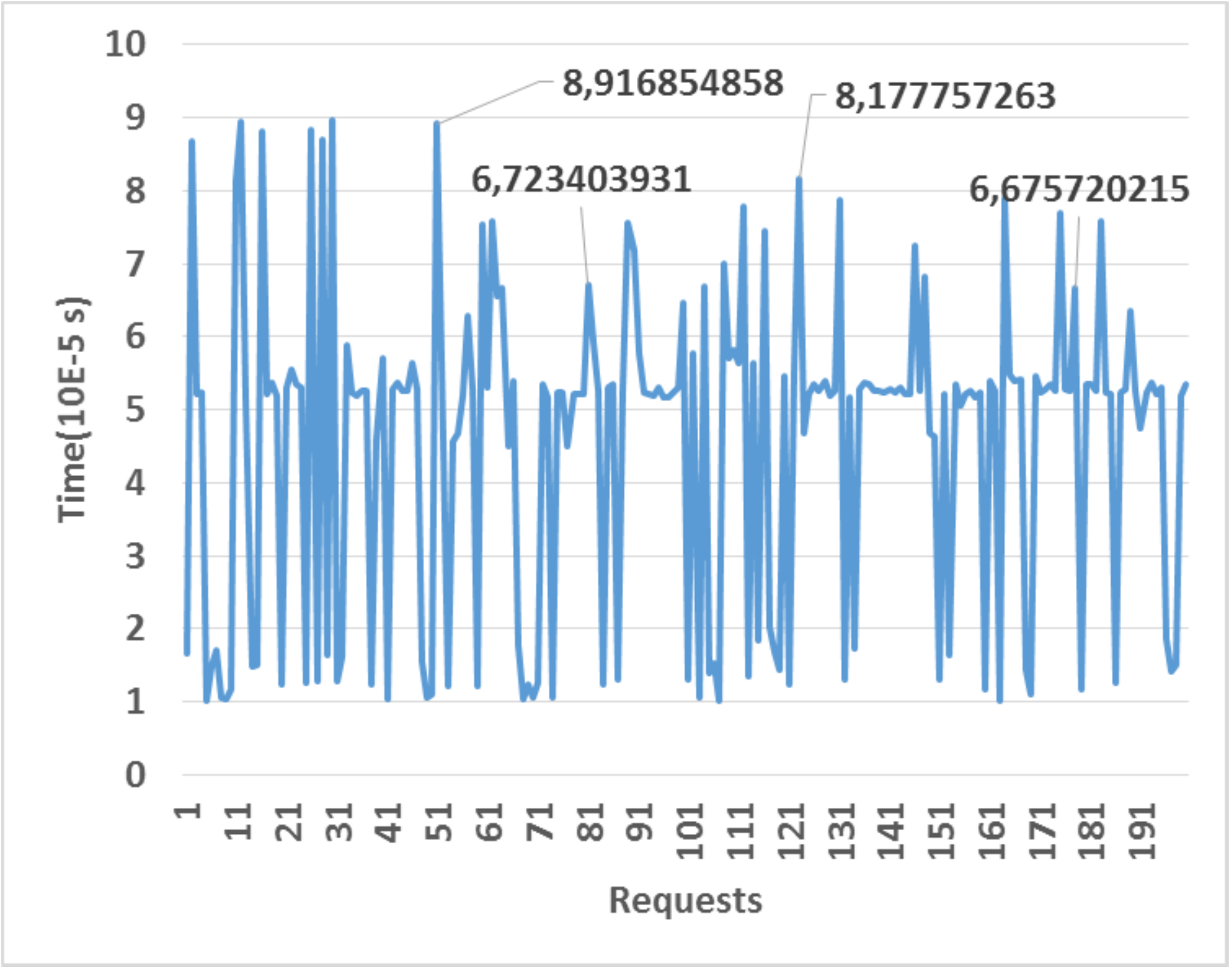}
        \caption{MOT16$-14$}
        \label{fig:off_search14}
    \end{subfigure}
    \caption{Offline search time}
    \label{fig:offline_search}
\end{figure*}

\begin{figure*}[h]
    \centering
    \begin{subfigure}[b]{0.49\textwidth}
        \centering
        \includegraphics[width=\textwidth]{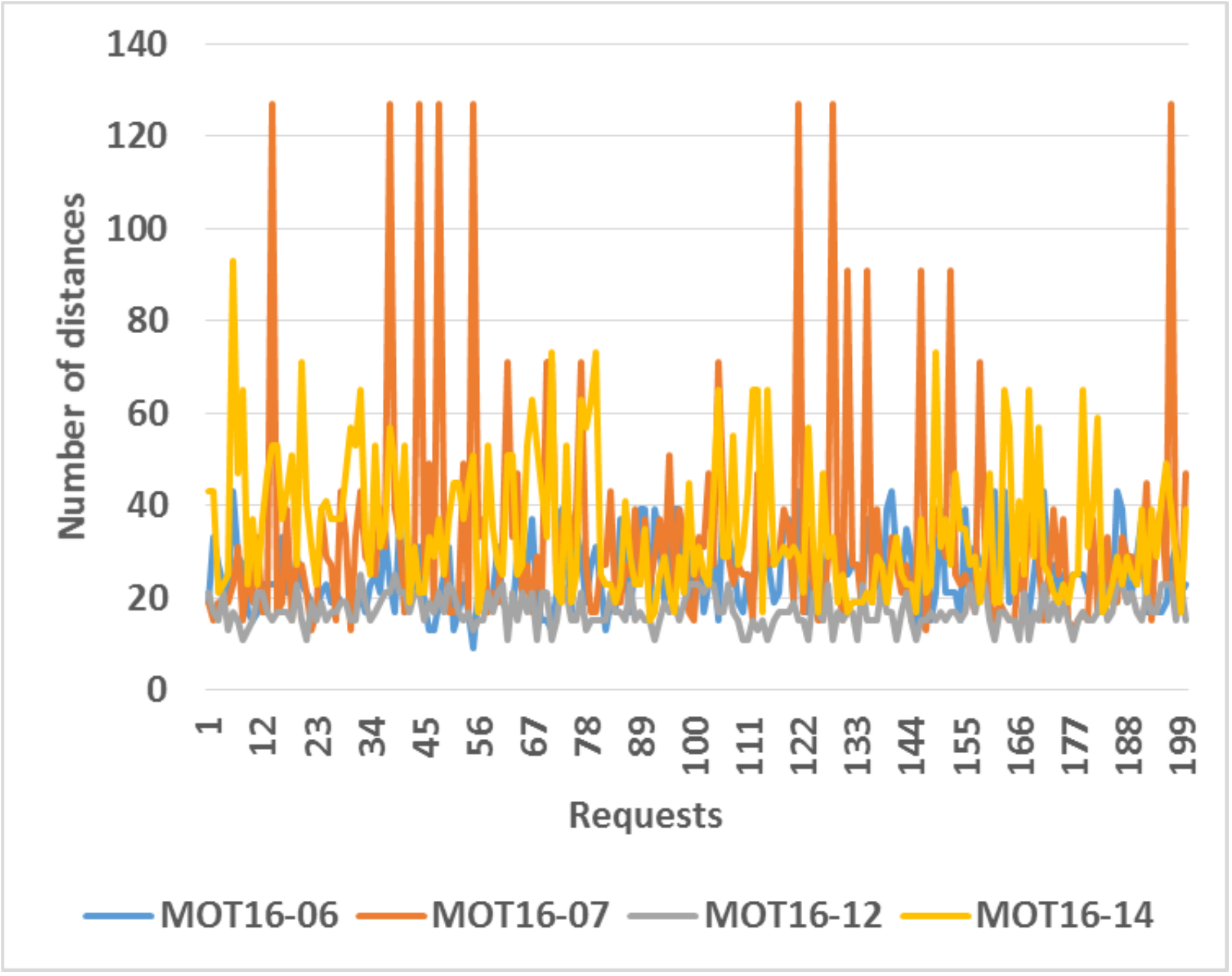}
        \caption{Number of distances}
        \label{fig:dist_offline}
    \end{subfigure}
    \hfill
    \begin{subfigure}[b]{0.49\textwidth}
        \centering
        \includegraphics[width=\textwidth]{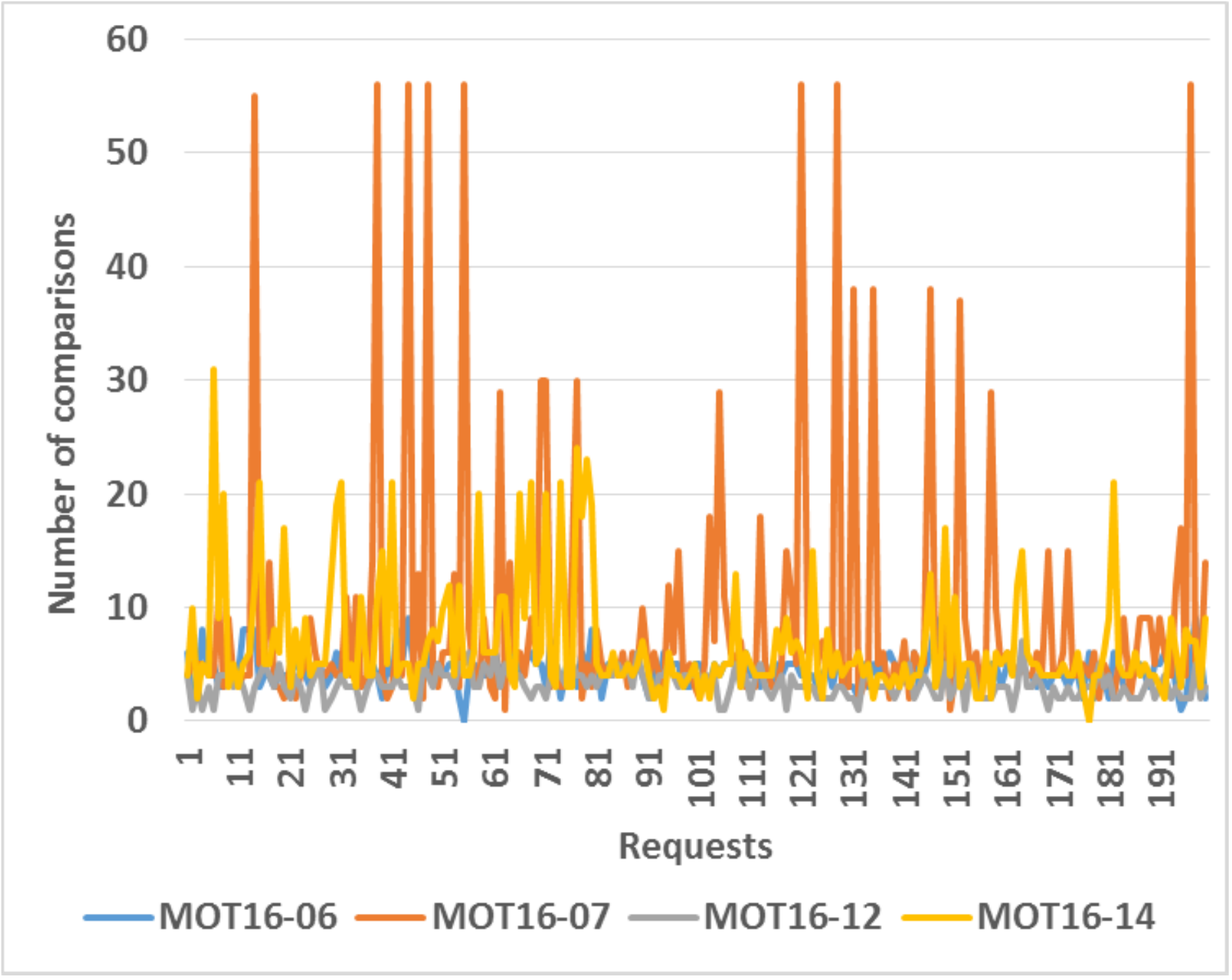}
        \caption{Number of comparisons}
        \label{fig:componline2}
    \end{subfigure}
    \hfill
    \caption{Offline search cost}
    \label{fig:dist_comp_off}
\end{figure*}

This section presents the performance of the proposed structure for offline search. The exact results are presented in Figures \ref{fig:offline_search} and \ref{fig:dist_comp_off}, which respectively demonstrate the time and cost of search processes (number of distances and comparisons). To better interpret the obtained results, we summarized the results by averaging the values obtained after performing 200 queries for each obtained tree (see Table \ref{tab:recherche_offline}).

{
\begin{table}
    \caption{Offline average search results}
    \label{tab:recherche_offline}
    \centering
    \begin{tabular}{lcccc}

        \hline
                      & MOT16$-06$ & MOT16$-07$ & MOT16$-12$ & MOT16$-14$ \\ \hline
        \#Distance    & 25         & 33         & 17         & 25         \\
        \#Comparaison & 4          & 9          & 3          & 7          \\
        Time (10E-5s) & 8.64       & 6.89       & 5.85       & 8.43       \\ \hline
    \end{tabular}

\end{table}
}
As shown in the results, the maximum execution time of the offline search varies in the range [$8.94\times10^{-5}s$, $9.98\times10^{-5}s$], which means that the most similar object found is located at a maximum depth level of the indexing structure (worst case). Also, we find that the minimum execution time to perform the offline search is 0.01$\times10^{-5}s$, which indicates that the object most similar to the search query is close to the root (best case). Regarding the average search time, Table \ref{tab:recherche_offline} shows that the time varies in the range [5.85$\times10^{-5}s$, 8.64$\times10^{-5}s$]. Concerning the search cost (number of distances and comparisons), we notice that it varies in the same way as the search time, as shown in Figure \ref{fig:dist_comp_off} and Figure \ref{fig:offline_search}.


\subsection{Evaluation of the Adaptive BCCF-tree based tracking system} \label{sec:systeme-eva}
We evaluated the performance of our solution in a multiple object tracking system proposed in \cite{wojke2017simple} on the MOT16 benchmark \cite{milan2016mot16}. This benchmark evaluates the performance of the tracking system on seven complex test sequences, including front-view scenes with a moving camera and a large collection of data sets. The evaluation is performed according to the standard metrics presented in Table \ref{tab:metrique}.
{
\begin{table}
    \caption{Tracking system evaluation metrics \cite{milan2016mot16} }
    \label{tab:metrique}
    \centering
    \begin{tabular}{p{2.5cm}cp{6.5cm}}

        \hline
        Metric                          & Abbrev             & Description                                                                                                                              \\ \hline

        Identity switches               & ID-Sw $\downarrow$ & Number of identity changes compared to the ground-truth                                                                                  \\

        Fragmentation                   & Frag $\downarrow$  & The total number of times a trajectory is fragmented (i.e., interrupted during tracking)                                                 \\
        Mostly lost                     & ML $\downarrow$    & The ratio of ground-truth paths that are covered by a track hypothesis for no more than 20\% of their respective lifetimes               \\
        Mostly tracked                  & MT $\uparrow$      & The ratio of ground-truth verified trajectories that are covered by a runway assumption for at least 80 \% of their respective lifetimes \\
        False positives                 & FP $\downarrow$    & The total number of false positives
        \\
        False negatives                 & FN  $\downarrow$   & The total number of false negatives (missed targets)                                                                                     \\

        Multi-Object Tracking Accuracy  & MOTA  $\uparrow$   & Multi-object tracking precision (+/- indicates standard deviation over all sequences)                                                    \\
        Multi-object tracking precision & MOTP $\uparrow$    & Summary of overall tracking accuracy in terms of bounding box overlap between the ground-truth and the reported location.                \\ \hline
    \end{tabular}
\end{table}
}
The results of our evaluation are presented in Table \ref{tab:res_test}. They show that the proposed solution successfully reduces the number of identity changes (\texttt{ID-Sw}). Compared to the Deep SORT method, we observe that the \texttt{ID-Sw} is reduced from 781 to 686, which corresponds to a reduction of 13\% and a reduction from 1423 to 686 which corresponds to a reduction of about 52\% compared to the SORT method. However, the trajectory fragmentation (\texttt{Frag}) of the tracked objects increases slightly compared to Deep SORT. This increase is due to the number of frames skipped before the confirmation of the object's detection. The purpose of this skip is to extract a better metadata representation of the detected object.

Concerning the overall tracking accuracy (\texttt{MOTP}), the results show that it is improved compared to all the mentioned methods. In addition, Table \ref{tab:res_test} show that our solution significantly reduced the false positive (\texttt{FP}) rate compared to DeepSort, with a rate over 25\%. Overall, our proposed solution demonstrates its efficiency in decreasing search time and increasing tracking quality.

Although there was an improvement in overall performance, the number of tracked objects with more than 80\% of their lifetime (\texttt{MT}) did not improve. This means that the improvements in \texttt{FP} and \texttt{ID-Sw} were not sufficient to overcome the lifespan thresholds that are detailed for the tracking metrics. On the other hand, the number of tracked objects with less than 20\% of their lifetime (\texttt{ML}) is mainly impaired by the increase in the number of \texttt{Frag}.
{
\begin{table}
    \caption{MOT16 tracking results}
    \label{tab:res_test}
    \centering
    \begin{tabular}{p{2cm}cccccccc}
        \hline
                                         & MT $\uparrow$ & ML $\downarrow$ & FP $\downarrow$ & FN  $\downarrow$ & ID-Sw $\downarrow$ & Frag $\downarrow$ & MOTA $\uparrow$ & MOTP $\uparrow$ \\
        \hline
        EAMTT \cite{sanchez2016online}   & 19            & 34.9            & \textbf{4407}   & 81223            & 910                & \textbf{1321}     & 52.5            & 78.8            \\
        POI  \cite{yu2016poi}            & \textbf{34}   & 20.8            & 5061            & \textbf{55914}   & 805                & 3093              & \textbf{66.1}   & 79.8            \\
        SORT \cite{bewley2016simple}     & 25.4          & 22.7            & 8698            & 63245            & 1423               & 1835              & 61.4            & 79.1            \\
        Deep SORT \cite{wojke2017simple} & 32.8          & \textbf{18.2}   & 12852           & 56668            & 781                & 2008              & 61.4            & 79.1            \\

        Our solution                     & 19.6          & 25.8            & 9557$^{(*)}$    & 75967            & \textbf{686}       & 2295              & 52.72           & \textbf{79.77 } \\

        \hline
    \end{tabular}
\end{table}
}

\section{Conclusion}
\label{section6}
The tracking of moving objects is a crucial step in intelligent video surveillance systems. To accomplish this step, it is essential to differentiate each object's identities to keep the maximum amount of information about it. To do this, a unique label or identifier is assigned to each detected object, regardless of the area, the time of their appearance, or the detection camera. The conservation of the same label is a crucial challenge due to the linear complexity of the sequential search to extract the labels in data increasing with time, the number of objects, and the number of cameras in the network. This issue is particularly prevalent in large-scale real-time video surveillance networks when metadata becomes 'big metadata'. This paper introduces a novel automated multi-object labeling system based on an indexing technique for efficient real-time tracking. This mechanism organizes the massive metadata of objects extracted during the tracking into a tree-based indexing structure. The main advantages of this structure in a tracking system is its logarithmic search and retrieval complexity, which implicitly reduces the search response time, and its quality of research results, which ensure coherent labeling of the tracked objects. This paper discusses the effectiveness of the label search algorithms and the tracking quality compared to other recent tracking systems on real-world datasets. Our evaluation results show that our solution reduces the \texttt{ID-Sw} and increases the overall tracking accuracy. As a result, the quality of the tracking is also improved. Overall, our proposed solution demonstrates its efficiency in decreasing search time and increasing tracking quality.

\section*{Declarations}
\subsection*{Conflict of interest/Competing interests}
The authors declare that they have no conflict of interest.

\subsection*{Funding}
No funding was received to assist with the preparation of this manuscript.

\bibliography{sn-bibliography}


\end{document}